\documentclass{article}

\usepackage{arxiv}
\usepackage[T1]{fontenc}    
\usepackage[pagebackref=true,breaklinks=true,colorlinks,bookmarks=false]{hyperref}
\usepackage{url}            
\usepackage{booktabs}       
\usepackage{amsfonts}       
\usepackage{nicefrac}       
\usepackage{microtype}      
\usepackage{graphicx}
\usepackage{textcomp}
\usepackage{rotating}

\usepackage{MnSymbol}
\usepackage[utf8]{inputenc}
\newtheorem{definition}{Definition}
\renewcommand{\thesection}{\arabic{section}}

\usepackage{parskip}
\usepackage[nopar]{lipsum}

\usepackage[utf8]{inputenc}
\usepackage[english]{babel}

\newtheorem{property}{Property}[definition]

\usepackage{titlesec}

\usepackage{subfig}
\usepackage{graphicx}
\usepackage[export]{adjustbox}
\usepackage{array}
\usepackage{adjustbox}
\usepackage{amsmath}
\usepackage{multirow}

\usepackage[ruled]{algorithm2e}
\usepackage{algorithmic}

\usepackage{pgfplotstable}
\usepackage{pgfplots}
\pgfplotsset{compat=1.15}
\usepackage{caption} 
\usepackage{floatrow}
\usepackage{cite}

\usepackage{ulem}

\title{\#PraCegoVer: A Large Dataset for Image Captioning in Portuguese}

\author{
  Gabriel Oliveira dos Santos, Esther Luna Colombini, Sandra~Avila \\
  Institute of Computing, University of Campinas (Unicamp), Brazil\\
  \texttt{g197460@dac.unicamp.br, \{esther,sandra\}@ic.unicamp.br}
 }

\begin{document}

\newcommand{\SA}[1]{{\color{blue}#1}}
\newcommand{\EC}[1]{{\color{red}#1}}
\newcommand{\GO}[1]{{\color{orange}#1}}
\newcommand{\Lembrete}[1]{{\color{cyan}#1}}

\tolerance=999
\sloppy

\maketitle

\begin{abstract}
Automatically describing images using natural sentences is an essential task to visually impaired people's inclusion on the Internet. 
It is still a big challenge that requires understanding the relation of the objects present in the image, their attributes, and the actions they are involved in. Then, visual interpretation methods are needed, but linguistic models are also necessary to verbally describe the semantic relations.
This problem is known as \textit{Image Captioning}. There are many datasets in the literature, but most contain only English captions, whereas datasets with captions described in other languages are scarce.
Recently, a movement called PraCegoVer arose on the Internet, stimulating users from social media to publish images, tag \textit{\#PraCegoVer} and add a short description of their content.
Thus, inspired by this movement, we introduce the \textit{\#PraCegoVer}, a multi-modal dataset with Portuguese captions based on posts from Instagram. It is the first large dataset for image captioning in Portuguese with freely annotated images.
Further, the captions in our dataset bring additional challenges to the problem: first, in contrast to popular datasets such as MS COCO Captions, \textit{\#PraCegoVer} has only one reference to each image; also, both mean and variance of our reference sentence length are significantly greater than those in the MS COCO Captions. These two characteristics make our dataset challenging due to its linguistic aspect and challenges to the image captioning problem. 
We publicly-share the 
dataset at \url{https://github.com/gabrielsantosrv/PraCegoVer}.
\end{abstract}


\section{Introduction}
\label{sec:introduction}

The Internet is becoming increasingly accessible, reaching a wide variety of audiences. However, little progress has been made in including people with disabilities. The scenario is even worse for visually impaired people since a significant part of the Internet content is exclusively visual, for instance, photos and advertising images. Although there are screen readers well evolved, they are still mostly dependant on some annotations added to the source code of websites, which in turn, in general, are not that descriptive. 

In light of this situation, in 2012, PraCegoVer \cite{criadoraPracegover} arose as a social movement, idealized by Patrícia Braille, that stands for the inclusion of people with visual impairments besides it has an educational propose. The initiative aims to call attention to the accessibility question. It stimulates users to post images tagged with \textit{\#PraCegoVer} and add a short description of their content. This project has inspired many local laws that establish that all posts made by public agencies on social media must refer to \textit{\#PraCegoVer} and contain a short description of the image.

Audio description is a type of transcription of the visual content into text, complying with accessibility criteria. Initially, it was thought to the audience with visual impairments, but it also benefits people with dyslexia and attention deficit. This description format has to follow some guidelines to communicate effectively with the blind interlocutor. 
For instance, it shall indicate what kind of image the post refers to (\textit{e.g.}, photographs, cartoons, drawings), contextualize the scene before the most critical elements through short sentences.
Most importantly, it should avoid using adjectives because the interlocutor is the one that has to interpret the image from the given description.

Automatically describing image content using natural sentences is an important task to include people with visual impairments on the Internet, making it more inclusive and democratic. This task is known as \textit{image captioning}. However, it is still a big challenge that requires understanding the semantic relation of the objects present in the image, their attributes, and the actions they are involved in to generate descriptions in natural language. Thus, in addition to visual interpretation methods, linguistic models are also needed to verbalize the semantic relations.

Inspired by the PraCegoVer project, we have been creating \textit{\#PraCegoVer} dataset
, which is a multi-modal dataset with images and audio descriptions in Portuguese. As far as we know, this is the first dataset proposed for the Image Captioning problem with captions\footnote{Hereinafter, we use caption and description interchangeably.} in Portuguese. Moreover, since our dataset's descriptions are based on posts that tagged \textit{\#PraCegoVer}, they are addressed to visually impaired people, reflecting the visual content. Besides, our dataset contains captions in which the mean and variance in length, \textit{i.e.} the number of words, are greater than those in MS COCO Captions \cite{chen2015microsoft}, making the former more challenging than the~latter.

To cope with the image captioning in Portuguese, 
the main contributions in this work are:
\begin{enumerate}
    \item We create one of the largest datasets for image captioning with Portuguese descriptions;
    \item We develop a framework for post-collection from a hashtag on Instagram; 
    \item We propose an algorithm to cluster and remove post duplication based on visual and textual information;
    \item We show that AoANet \cite{AoANet_2019}, a state-of-art algorithm for MS COCO Captions, has a poor result on our dataset.
\end{enumerate}

We hope that \textit{\#PraCegoVer} dataset encourages more works addressing the automatic generation of descriptions in Portuguese. We also intend to contribute to the blind Portuguese speaker community. 

The remainder of the text is organized as follows. We survey the related work in Section~\ref{sec:related_works}. We detail the \textit{\#PraCegoVer} dataset in Sections~\ref{sub:data_collection}, \ref{sec:dup_detection}, \ref{sec:preprocessing}, \ref{sub:dataset_statistics}, including data collection, preprocessing, analysis, and statistics.  We describe the experiments  with a discussion of the main findings in Section~\ref{sec:experiments}.  We conclude the paper in Section~\ref{sub:conclusion}. We also provide a datasheet for the \textit{\#PraCegoVer} dataset, as proposed by Gebru \textit{et al.} \cite{gebru2020datasheets}, in \ref{app:datasheets}.

\section{Related Work}
\label{sec:related_works}

The image captioning task has been accelerated thanks to the availability of a large amount of annotated data in relevant datasets, for instance, Flickr8k~\cite{Flickr8K2013}, Flickr30k~\cite{FlicKr30K2017}, and MS COCO Captions \cite{chen2015microsoft}. 

Microsoft Common Objects in COntext (MS COCO) Captions is a dataset created from the images contained in MS COCO \cite{MSCoco2014} and human-generated captions. MS COCO Captions dataset comprises more than 160k images collected from Flickr, distributed over 80 object categories, with five captions per image. Its captions are annotated by human annotators using the crowdsourcing platform Amazon Mechanical Turk (AMT). The annotators were told, among other instructions, not to give people proper names and write sentences with at least eight words. As a result, the descriptions' average sentence length is about 10 words and with no proper names.

Many large-scale datasets have been created \cite{rashtchian2010collecting, farhadi2010every, elliott2013image, zitnick2013learning, kong2014you, harwath2015deep, gan2017stylenet, krishna2017visual, sidorov2020textcaps}, but in contrast to the previous ones, they employ automated pipelines. One example of a dataset that follows this approach is the Conceptual Captions dataset~\cite{sharma2018conceptual} which has more than 3.3M pairs of images and English captions. It was created by crawling web pages and extracting images and the alt-text HTML attribute associated with them. Images and captions are automatically filtered as well as cleaned aiming to select informative and learnable data.

To explore real-world images, Agrawal \textit{et al.} proposed nocaps \cite{agrawal2019nocaps}, a benchmark that consists of validation and test set with 4,500 and 10,600 images, respectively, annotated with 11 human-generated captions per image. This dataset is created by filtering images from the Open Images V4 dataset \cite{krasin2017openimages} and selecting images based on their object categories. Moreover, nocaps has more objects per image than MS COCO, and it has 600 object categories, whereas MS COCO has only 80. This benchmark evaluates models on how well they generate captions for objects not present in the dataset on which they are trained.

Recently, Gurari \textit{et al.} proposed VizWiz-Captions dataset \cite{gurari2020captioning} focused on the real use case of the models by blind people. It represents a paradigm shift of image captioning towards goal-oriented captions, where captions faithfully describe a scene from everyday life and answer specific needs that blind people might have while executing particular tasks. This dataset consists of 39,181 images taken by people who are blind, each image paired with five captions annotated by using the AMT platform. They also have metadata that indicates whether a text is present on the image and the image quality issues. The overlap between VizWiz-Captions and MS COCO content is about 54\%, which means a significant domain shift in the content of pictures taken by blind photographers and what artificially constructed datasets represent.

InstaPIC-1.1M~\cite{instaPic_chunseong2017attend} was created by collecting posts from Instagram, comprising 721,176 pairs of image-caption from 4.8k users. Based on the 270 selected hashtags, they crawled the Instagram APIs to filter the posts and collect the images and captions. The major problem in the InstaPIC-1.1M dataset is that the captions may not reflect the image content because they are based on what Instagram users write about their posts, which can be quite vague and do not describe the visual content. For example, ``Autumn is the best.'' and ``We take our adventure super seriously. \#selfiesunday'' are vague captions present in its training~set. 

Our approach also is based on Instagram posts, but in contrast to InstaPIC-1.1M and Conceptual Captions, we collect only captions where \textit{\#PraCegoVer} is tagged. Moreover, we clean the captions maintaining just the audio description part written by supporters of the PraCegoVer movement. Thus, similarly to VizWiz-Captions, our dataset's descriptions are addressed to visually impaired people and reflect the visual content. Still, our dataset contains captions with 40 words on average, while those in MS COCO Captions have only ten words, and the variance of sentence length in our dataset is also more significant. On the other hand, since our dataset is freely annotated, the descriptions might contain proper names that can be removed, and we consider this an essential point for future improvements. Finally, \textit{\#PraCegoVer} is the only dataset, as far as we know, that comprises Portuguese descriptions, in contrast to the others that comprehend English captions. Also, the average sentence length the variance of length in terms of the number of words make \textit{\#PraCegoVer} a challenging~dataset. 
\section{Data Collection}
\label{sub:data_collection}

Several companies and government agencies have joined the campaign \textit{\#Pra\-CegoVer}, thereby posting images, including their audio description, on social networks such as Facebook, Instagram, and Twitter. Although these data are available on the internet, they are not easy to collect because the platforms generally limit their public APIs (Application Programming Interface). Moreover, the restrictions may vary among the platforms; for instance, Facebook provides an API that only allows access to posts from one user, making it challenging to collect data on a large scale. On the other hand, Instagram permits access to public posts from many users, limiting the search to posts published in the last seven days.

In this work, we have collected data only from Instagram since it focuses on image sharing and allows us to filter posts by hashtag. As mentioned before, Instagram limits the filter by hashtag to posts published in the last seven days, therefore to overcome this obstacle, we first search for posts related to the hashtag \textit{\#PraCegoVer} and save just the profiles. In the following step, we visit these profiles looking for more posts tagged with the hashtag we are interested in. Inside the profile pages, there are no limitations regarding date or quantity. Thus, we have access to all images published by that user because they are~public, and finally we can collect the posts.

\begin{figure}[htp!]
    \includegraphics[width=\textwidth]{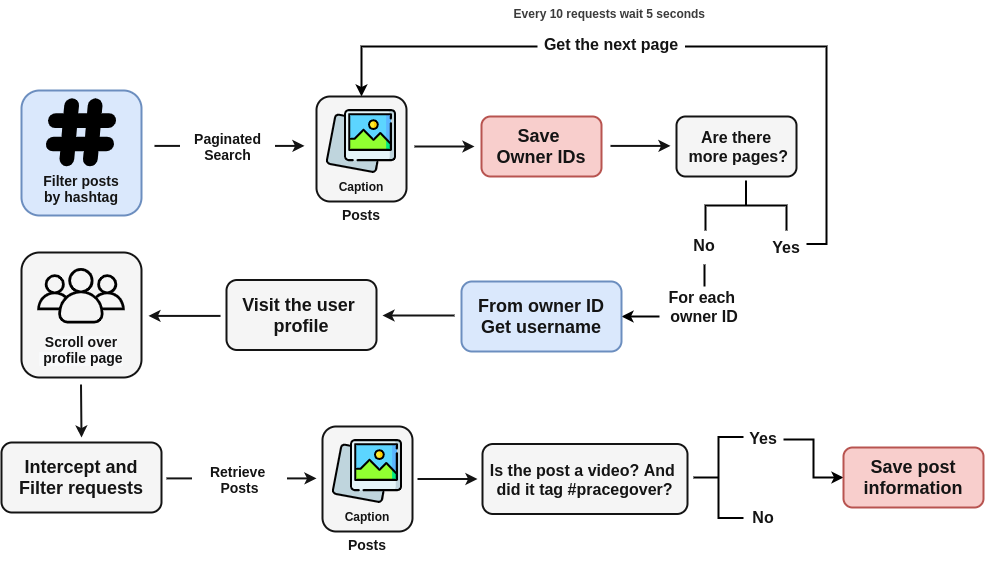}
    \caption{Diagram illustrating the pipeline of data collection. We start filtering the posts by hashtags and save the profile identifiers. From these identifiers, we retrieve the usernames, which are used to visit each profile page. We use Selenium Wire 
    to automate the browser to visit all the profile pages and scroll over them retrieving the posts. Icons made by Freepik and prosymbols from ``www.flaticon.com".}
    \label{fig:pipeline_collection}
    \centering
\end{figure}

We execute this process daily and incrementally, storing: images, their width and height, their captions, post identifiers, post owners, post date, and the collection date. In this way, we can collect posts published any time ago, instead of up to the past seven days as restricted in the hashtag page.  We highlight that we ensure that the robot never accesses posts from private profiles, which would require an acceptance from each account's owner.
However, there may be profiles that became private after we had collected their posts. Figure~\ref{fig:pipeline_collection} illustrates our pipeline of data collection.

\section{Duplication Detection and Clustering}
\label{sec:dup_detection}

We collected our data from Instagram, where people usually share similar posts multiple times, by just changing a few details of images such as adding a new logotype, cropping and rotating the image, changing color scale, and so forth, as illustrated in Figure \ref{fig:cluster_perfum}. We consider these images as duplications because they do not bring new information to the dataset. Thus, removing such items from the dataset is essential because models may get overfitted in those duplicated examples. Also, there are posts in which the images are duplicated but not the caption, and we can use these different captions to create a structure similar to MS COCO Captions, where one image is associated with multiple captions. This section describes our proposed algorithm to identify and cluster post duplications. This algorithm uses image features to cluster similar posts and leverages the textual information to eliminate eventual ambiguity.

\subsection{Duplications}
\label{sub:duplications}

The concept of duplication depends on the application and the data. Then, it is important to define what we consider duplicated posts. Formally, let a post be a tuple of image and caption, $post = (image, caption)$, and $dist(\cdot,\cdot)$ be a distance function, then we define:

\begin{definition}[Image Duplication]
\label{def:image_similarity}
  Given two images $image_1$ and $image_2$, they are duplicated if $dist(image_1, image_2) \leq t_{img}$, for some predefined threshold $t_{img}$. We denote this duplication by $image_1 \sim image_2$.
\end{definition}

\begin{definition}[Caption Duplication]
\label{def:caption_similarity}
  Given two captions $caption_1$ and $cap\-tion_2$, they are duplicated if $dist(caption_1, caption_2) \leq t_{cpt}$, for some predefined threshold $t_{cpt}$. We denote this duplication by $caption_1 \sim caption_2$.
\end{definition}

\begin{definition}[Post Duplication]
\label{def:post_dup}
    Given two posts $post_1 = (image_1, caption_1)$ and $post_2 = (image_2, caption_2)$, they are considered as a duplication if, and only if $image_1 \sim image_2$ and $caption_1 \sim caption_2$. We denote this duplication by $post_1 \sim post_2$.
\end{definition}

From this definition, we have the transitivity property that is the basis for our algorithm to cluster duplications. 
\begin{property}[Transitivity]
\label{prop:img_transitivity}
Given the posts $post_1$, $post_2$ and $post_3$, then $post_1 \sim post_2$ and  $post_2 \sim post_3$ $\Rightarrow$  $post_1 \sim post_3$.
\end{property}

\begin{figure}[ht!]
    \centering
    \subfloat[User 1. Caption: ``Na imagem, o Zaad Mondo aparece sob uma superfície tomada completamente por juníperos. São esses frutos tão pequenos que compõem o seu acorde tão marcante.'']{{\includegraphics[scale = 0.125]{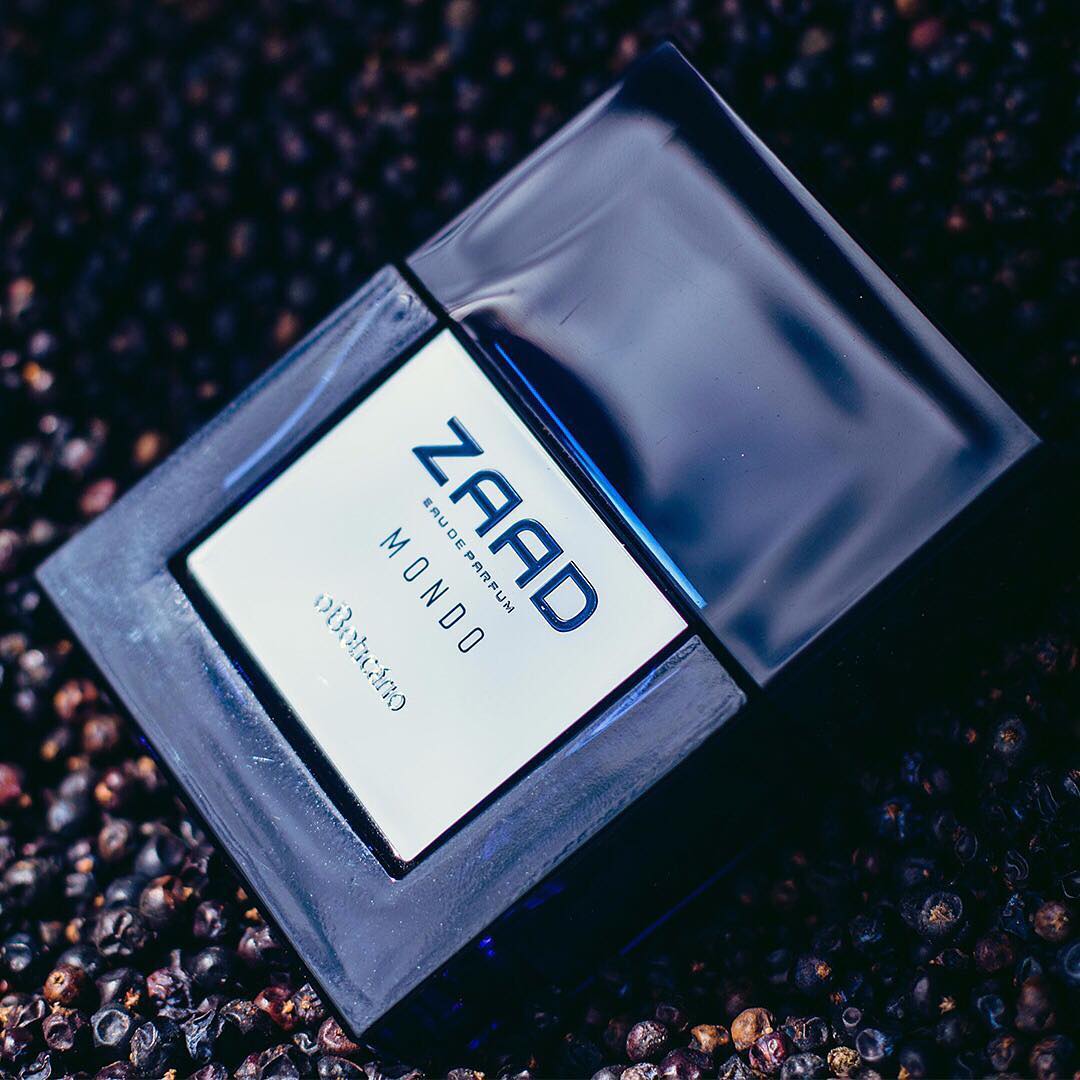} }}
    \qquad
    \subfloat[User 2. Caption: ``Na imagem, o Zaad Mondo aparece sob uma superfície tomada completamente por juníperos. São esses frutos tão pequenos que compõem o seu acorde tão marcante.'']{{\includegraphics[scale = 0.125]{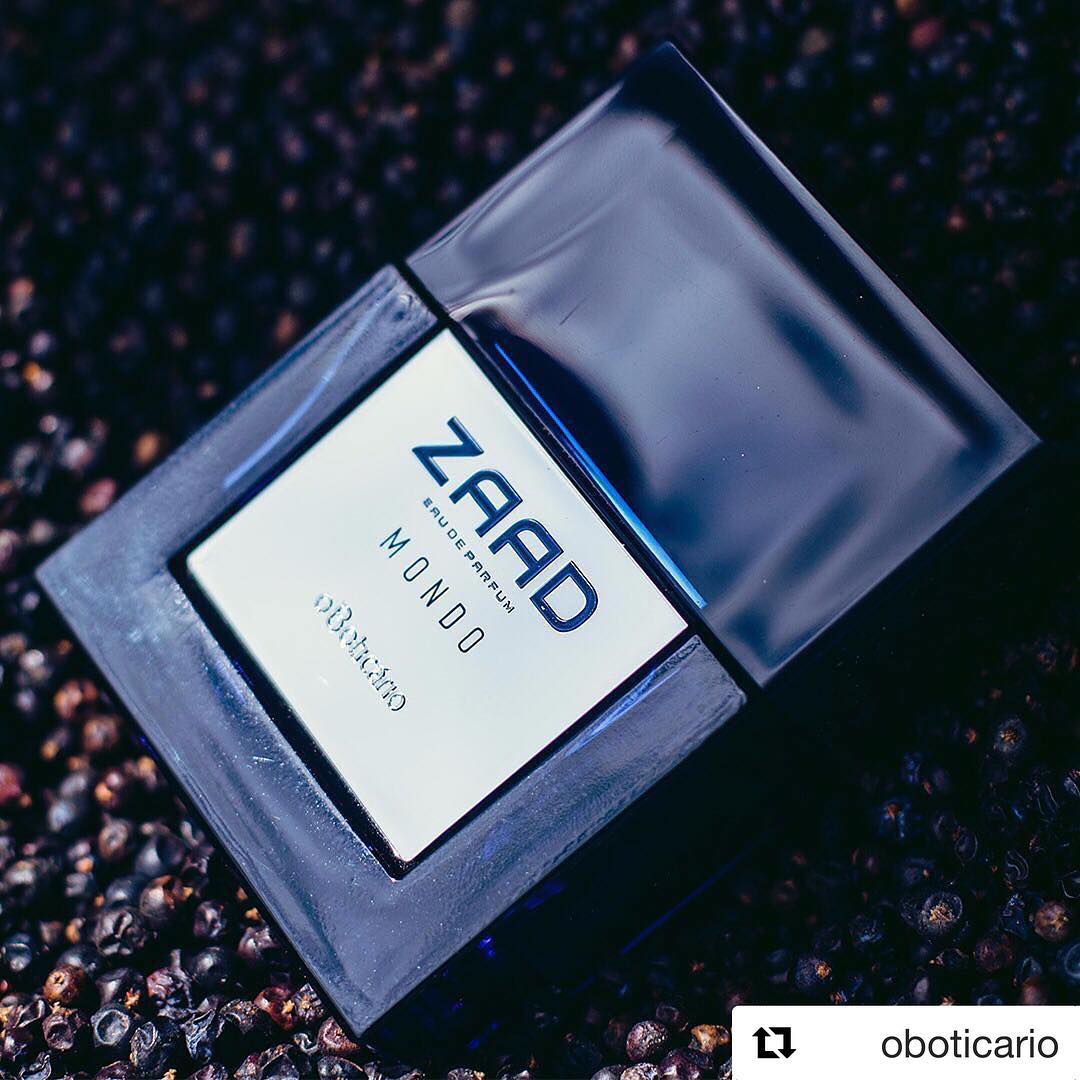} }}
   \caption{Two similar images posted on Instagram by two different profiles: User 1 and User~2. It can be seen that image (b) is similar to image (a), however, it contains a logo in the lower right corner that is not present in image (a). Moreover, both posts have the same caption, thus we consider them as a duplication.}
    \label{fig:dup_images}
\end{figure}

Figure~\ref{fig:dup_images} illustrates an example of two posts considered duplicated. They have similar images and texts.

\subsection{Duplication Clustering}
\label{sub:dup_grouping}

Clustering the duplicated posts is an essential step of the dataset creation process because these duplications may lead to unwanted bias. Also, we must avoid similar instances in the train and test sets to guarantee that the results obtained by algorithms trained on this dataset are representative.

We have designed the clustering Algorithm~\ref{algo:grouping_dup} based on Definition~\ref{def:post_dup} and Transitivity Property~\ref{prop:img_transitivity}. On this subject, we create a similarity graph that keeps the duplicated posts in the same connected component. Formally, the similarity graph is an undirected graph $G(V, E)$ such that for all $post_i$ we create a vertex $v_i$, and for each pair of vertices $v_i, v_j \in V$, there exist the edge $(v_i, v_j) \in E$ if, and only if $post_i \sim post_j$. Note that from the Transitivity Property if there exists a path $(v_i, e_i, v_{i+1}, e_{i+1}, .., v_j)$, then $post_i \sim post_{i+1} \sim ... \sim post_j$. Thus, it is guaranteed that all duplicated posts will be kept in the same connected component of similarity graph. Algorithm \ref{algo:grouping_dup} is the pseudocode of our clustering method.

\begin{algorithm}[htp]
\begin{algorithmic}[1]
    \REQUIRE number of posts $n$, distance matrices $D_{cpt}$, $D_{img}$, and thresholds $t_{cpt}$, $t_{img}$
    
    \STATE $graph_{sim}[i][j] \leftarrow 0 \forall i,j \in \{1, .. , n\}$

    \FORALL{$i \in [1 .. n]$}
        \FORALL{$j \in [i+1 .. n]$}
            \IF{$D_{img}[i][j] \le t_{img}$ and $D_{cpt}[i][j] \le t_{cpt}$}
                \STATE $graph_{sim}[i][j] \leftarrow 1$
                \STATE $graph_{sim}[j][i] \leftarrow 1$
            \ENDIF
        \ENDFOR
    \ENDFOR
    
    \STATE $visited \leftarrow \emptyset$
    \STATE $clusters \leftarrow \emptyset$
    
    \FORALL{$v \in [1 .. n]$}
        \IF{$v \not\in visited$}
            \STATE $dups \leftarrow \emptyset$
            \STATE $DFS(graph_{sim}, n, v, visited, dups)$
            \STATE $clusters.append(dups)$
        \ENDIF
    \ENDFOR
        
    \STATE \textbf{return} $cluster$ \COMMENT{A list with sets of duplications clustered.}
    
\end{algorithmic}    
\caption{Clustering duplications}
\label{algo:grouping_dup}
\end{algorithm}

\begin{algorithm}[htp!]
\begin{algorithmic}[1]
    \REQUIRE $graph_{sim}, n, v, visited, dups$

    \STATE $visited[v] \leftarrow True$
    \STATE $dups.append(v)$

    \FORALL{$u \in [1 .. n]$}
        \IF{$graph_{sim}[v][u] = 1$ and not\ $visited[u]$}
            \STATE $DFS(graph_{sim}, n, u, visited, dups)$
        \ENDIF
    \ENDFOR
\end{algorithmic}    
\caption{DFS}
\label{algo:dfs}
\end{algorithm}

Algorithm \ref{algo:grouping_dup} requires the distance matrices $D_{cpt}$ and $D_{img}$ of captions and images, respectively. Also, it requires caption and image thresholds, denoted by $t_{cpt}$ and $t_{img}$
. The distance matrices can be constructed using any distance metric. We found empirically that the cosine distance is a good metric for both images and captions. Moreover, we conducted a grid search varying the thresholds $t_{img}$ and $t_{cpt}$ in the set $\{0.02, 0.05, 0.10, 0.20\}$. We concluded that low thresholds result in many small clusters of duplicated posts, while high thresholds result in few large clusters with some miss clustered posts. In our experiments, we obtained the best clusters when we set the thresholds to $t_{img} = t_{cpt} = 0.10$.
 
\begin{figure}[htp]
    \centering
    \subfloat[]{\includegraphics[scale=0.35]{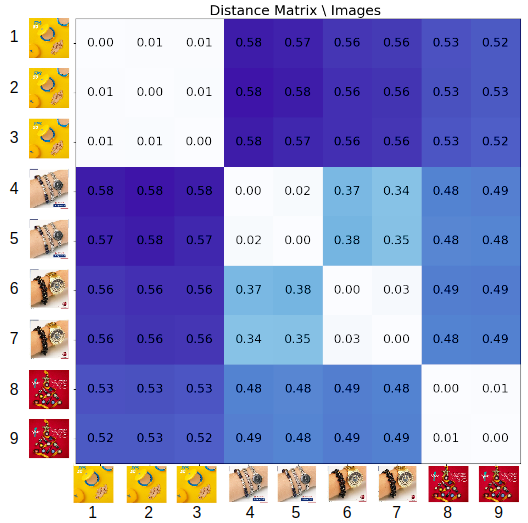}\label{fig:image_dist_matrix}}\hspace{0.05cm}
    \subfloat[]{\includegraphics[scale=0.35, trim=0 0.275cm 0 0, clip ]{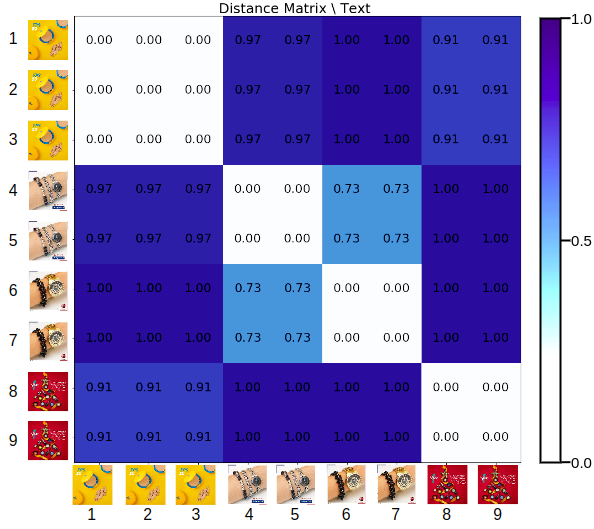}\label{fig:text_dist_matrix}}
    \caption{Distance matrix constructed from the pair-wise cosine distance based on (a) image features, and (b) text features.}
\end{figure}

\begin{figure}[htp]
    \centering
    \subfloat[Similarity graph based only on image distances.]{{\includegraphics[scale = 0.35]{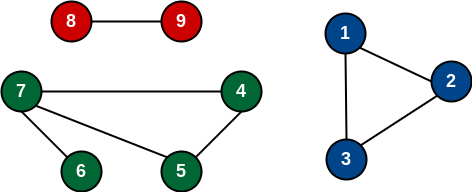}\label{fig:similarity_graph_a} }}
    \qquad
    \subfloat[Similarity graph based only on both image and text distances.]{{\includegraphics[scale = 0.35]{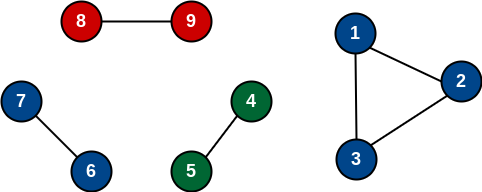}\label{fig:similarity_graph_b} }}
  
   \caption{Similarity graphs considering only image features and considering both visual and textual features. It can be seen that when just image features are taken into account, the algorithm return the clusters \{1,2,3\}, \{4,5,6,7\} and \{8,9\}, because the distances are lower than or equal to $t_{img} = 0.35$. However, when we also consider the textual information, it returns the clusters \{1,2,3\}, \{4,5\}, \{6,7\} and \{8,9\}.}
    \label{fig:similarity_graph}
\end{figure}

To illustrate this algorithm, let us consider the nine following posts, and suppose the distance matrices with respect to images and texts are shown in Figure~\ref{fig:image_dist_matrix} and Figure~\ref{fig:text_dist_matrix}, respectively. Also, consider the thresholds $t_{img} = 0.35$ and $t_{cpt} = 0.10$. It can be seen that if only consider the image distance, then we will have the graph in Figure~\ref{fig:similarity_graph_a}, where each vertex represents a post and the clusters \{1,2,3\}, \{4,5,6,7\} and \{8,9\} represent connected components, because the distances among posts in the same cluster are lower than or equal to $t_{img} = 0.35$. However, when we also leverage the textual information, the cluster \{4,5,6,7\} is split into \{4,5\} and \{6,7\}, as illustrated in Figure~\ref{fig:similarity_graph_b}, because the text distance among the captions of posts in these clusters is lower than or equal to $t_{cpt} = 0.10$. Therefore, textual information can improve the identification of duplications by enhancing the context.
  
\section{Preprocessing and Data Analysis}
\label{sec:preprocessing}

In this section, we describe the text and image preprocessing necessary to execute Algorithm \ref{algo:grouping_dup} that clusters post duplications and to split the final dataset into a training, validation, and test sets. 

\begin{figure}[htp!]
    \includegraphics[width=0.75\textwidth]{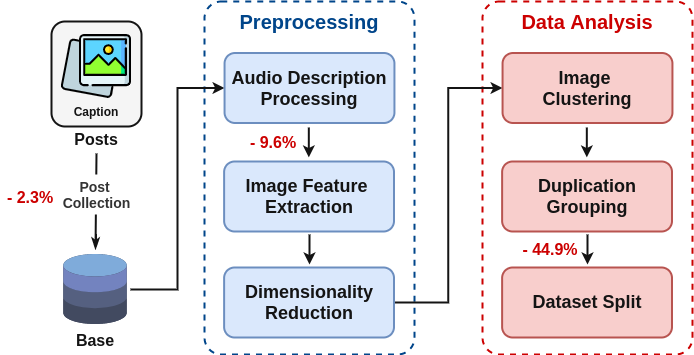}
    \caption{Overview of the whole pipeline from the data collection to the dataset split. First, we collect the data, clean the captions to obtain the audio descriptions, and extract the image and text features. Finally, we analyze the data to remove duplications and split the dataset into training, validation, and test sets. We highlight the percentage of posts lost in each step. Icons made by Freepik and Smashicons from ``www.flaticon.com''.}
    \label{fig:pipeline_dataset}
    \centering
\end{figure}

Figure~\ref{fig:pipeline_dataset} illustrates our pipeline and highlights the percentage of posts lost in each step concerning the total amount of collected posts. We split our pipeline into two phases: preprocessing and data analysis. The \textit{preprocessing} phase consists of the general processing of texts and images. On the other hand, the steps in the \textit{data analysis} phase are not only used as preprocessing but are also used to explore the data by visualizing the clusters and duplications.

The first step of our pipeline comprehends collecting the data from Instagram (see Section \ref{sub:data_collection}). Then, we clean the captions to obtain the descriptions, and we extract images and text features. Next, we reduce the dimensionality of image feature vectors to optimize processing, clustering the images to analyze the data and to remove duplicates. Finally, we split the dataset into training, validation, and test sets.

Moreover, we highlight in Figure~\ref{fig:pipeline_dataset} the loss in each step of the pipeline. About 2.3\% of posts are lost during the post-collection process because of profiles that become private during this process. Besides, 9.6\% of posts have malformed captions. They do not follow the main pattern, which consists of the ``\#pracegover'' followed by the description. Thus it is tough to extract the actual caption from the whole text. Therefore we remove them. Finally, 44.9\% of the total amount of posts have duplicated either caption or image, which may easily overfit the models, then we also remove these cases from the dataset. In total, about 56.8\% of data is lost or removed. It is worth noting that there is a loss of data inherent to the data source and the process.

\subsection{Audio Description Processing} 
\label{sub:caption_preprocessing}

Although the ones that joined the \textit{\#PraCegoVer} initiative produce a considerable amount of labeled data ``for free'', there are some inherent problems in the free labeling process. For instance, misspellings and the addition of emoticons, hashtags, and URL link marks in the captions. Furthermore, the captions often have some texts besides the audio description itself. Therefore, it is needed to preprocess the texts to extract the actual audio description part. 
We identified patterns in those texts by thoroughly reading many of them, and we used regular expressions to find the audio description within the texts. For instance, the audio description, in general, comes right after the hashtag \textit{\#PraCegoVer}, so we first crop the caption keeping just the text after this hashtag. Then, we use regular expressions to remove emoticons, hashtags, URL links, and profile marks, but it might lead to miss punctuations at the end of the texts that we also remove. For example, some posts have captions with a mark of ``end of audio descriptions''. Thus, we also use it as an end delimiter. Finally, we convert the texts into lower case, remove stopwords, and transform them into TF-IDF (Term Frequency–Inverse Document Frequency) vectors.

\begin{figure}[htp!]
    \includegraphics[width=0.95\textwidth, frame=0.5pt]{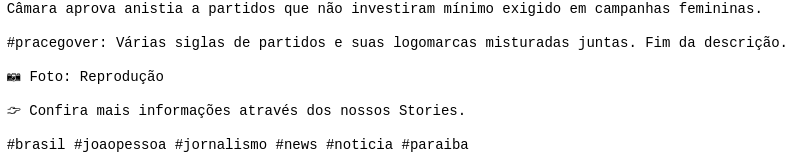}
    \caption{Example of a real caption in which is tagged the hashtag \textit{\#PraCegoVer}. After the extraction of audio description, we have the text ``Várias siglas de partidos e suas logomarcas misturadas juntas.''. Observe that there are emoticons and hashtags in the original caption, but we remove them.}
    \label{fig:audiodescription}
    \centering
\end{figure}

Figure~\ref{fig:audiodescription} shows a real example of the caption. After it is cleaned, the final text is ``Várias siglas de partidos e suas logomarcas misturadas juntas.''.\vspace{0.25cm}

\subsection{Image Feature Extraction}
\label{sub:image_feature_extraction}

Algorithm \ref{algo:grouping_dup} is inefficient in terms of memory consumption. Hence, it is unfeasible to compute two distance matrices of the whole dataset. We clustered posts using only the features extracted from images to tackle this problem so that duplicated images would be within the same cluster. Then, we run Algorithm~\ref{algo:grouping_dup} over each cluster, which reduces the search space and, as a result, requires less memory space. Nevertheless, a thorough cluster analysis is needed because the clusters also can be large and meaningless, which would lead to the previous problem. We extracted image features using MobileNetV2~\cite{mobilenetv2}, a convolutional neural network (CNN) and a popular choice for feature extractors.

\subsection{Dimensionality Reduction}
\label{sub:dimensionality_reduction}

\begin{figure}[htp!]
    \includegraphics[scale=0.55]{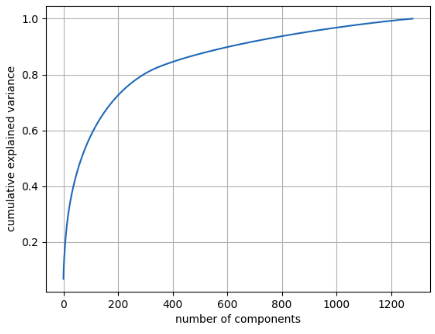}
    \caption{The graph shows the cumulative explained variance by the number of components computed using the algorithm of PCA. It can be observed that 900 dimensions account for 95\% of the total variance in the image features.}
    \label{fig:explained_variance}
    \centering
\end{figure}

MobileNetV2 returns image feature vectors with dimension 1280. Thus, to optimize memory usage, we decided to reduce dimensionality. We used Principal Component Analysis (PCA) \cite{tipping1999probabilistic} to estimate the number of dimensions needed to keep around 95\% of the explained variance (see Figure \ref{fig:explained_variance}), and we found 900~dimensions. Then, we reduce the dimensionality from 1280 to 900 dimensions by using UMAP (Uniform Manifold Approximation and Projection for Dimension Reduction) \cite{umap2018}, a non-linear dimension reduction method that preserves structures from the high-dimensional space into the lower dimensional embedding. We executed the algorithm considering 900 dimensions, a neighborhood of 80, the minimum distance between embedded points equals 0, and correlation as the metric.

\subsection{Image Clustering}
\label{sub:clustering_method}

We used HDBSCAN (Hierarchical Density-Based Spatial Clustering of Applications with Noise) \cite{hdbscan2017mcinnes} to cluster the reduced feature vectors. The hierarchical characteristic of this algorithm helps to find classes and subclasses. However, because of the nature of our data, which is sparse, the algorithm generates many outliers. To overcome this problem, we execute the algorithm iteratively: first, we cluster the data using HDBSCAN, and then we reapply this clustering algorithm to those data points assigned as outliers and repeat this process until it has reached a satisfactory amount of outliers. After we have the clustered images, we compute the distance matrices for captions and images only for those posts whose images belong to the same cluster. Thus, we significantly reduce memory usage. 

\subsection{Dataset Split}
\label{sub:dataset splits}


We thoroughly split \textit{\#PraCegoVer} dataset,  guaranteeing the quality of sets, and above all, the quality of the test set. We aim to avoid eventual bias related to profiles and duplications. Thus, we start by explicitly removing the detected duplications (see Section \ref{sec:dup_detection}). Then, we cluster the posts by owners and add all posts from a profile into the same set (either training, validation, or test), such that two different sets do not contain posts from the same profile. This way, we can test and validate the models in cross-profile sets. Finally, we split our dataset considering the proportion 60\% for training, 20\% for validation, and 20\% testing.

\section{Dataset Statistics}
\label{sub:dataset_statistics}

\begin{figure}[htp!]
    \includegraphics[width=0.7\textwidth]{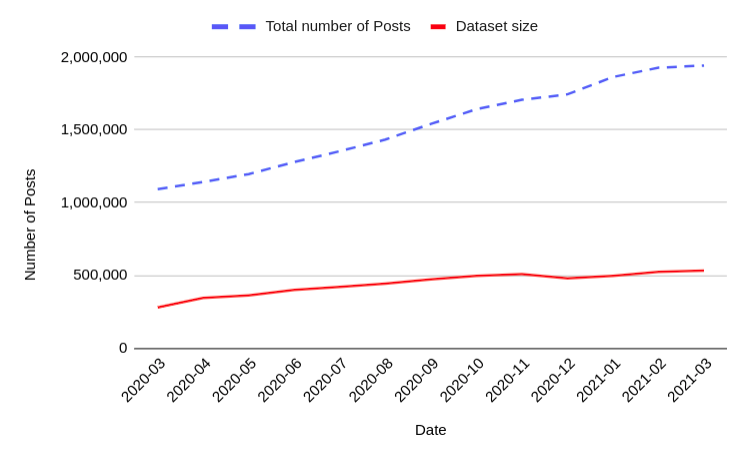}
    \caption{The total number of posts tagging \#PraCegoVer (dashed line) and  \textit{\#PraCegoVer} dataset size (solid line) throughout the time. 
    }
    \label{fig:dataset_growth}
    \centering
\end{figure}

We have collected more than 520,997 posts from 14,000 different profiles on Instagram. Nevertheless, after we clean the data, there will be only 45\% of the total remaining for training models, the other 55\% are removed throughout the pipeline as illustrated in Figure \ref{fig:pipeline_dataset}. Furthermore, our dataset contains a wide variety of images, which is essential to train models that generalize to real-world images on the Internet. Also, we highlight that the \textit{\#PraCegoVer} dataset is growing over time because we are continuously collecting more data. Figure \ref{fig:dataset_growth} shows the growth of the total number of posts tagging \#PraCegoVer (dashed line) and the rise of \textit{\#PraCegoVer} dataset size through time. We can see that the total amount of posts tagging \#PraCegoVer is greater than we can collect. It is due to Instagram mechanisms that limit our access to the posts. On average, our dataset has 30\% of the total amount of posts. Despite this fact, the  \textit{\#PraCegoVer} dataset still has enormous growth potential because the number of organizations joining the \#PraCegoVer movement is increasing.

\begin{figure}[tp!]
    \includegraphics[width=0.75\textwidth]{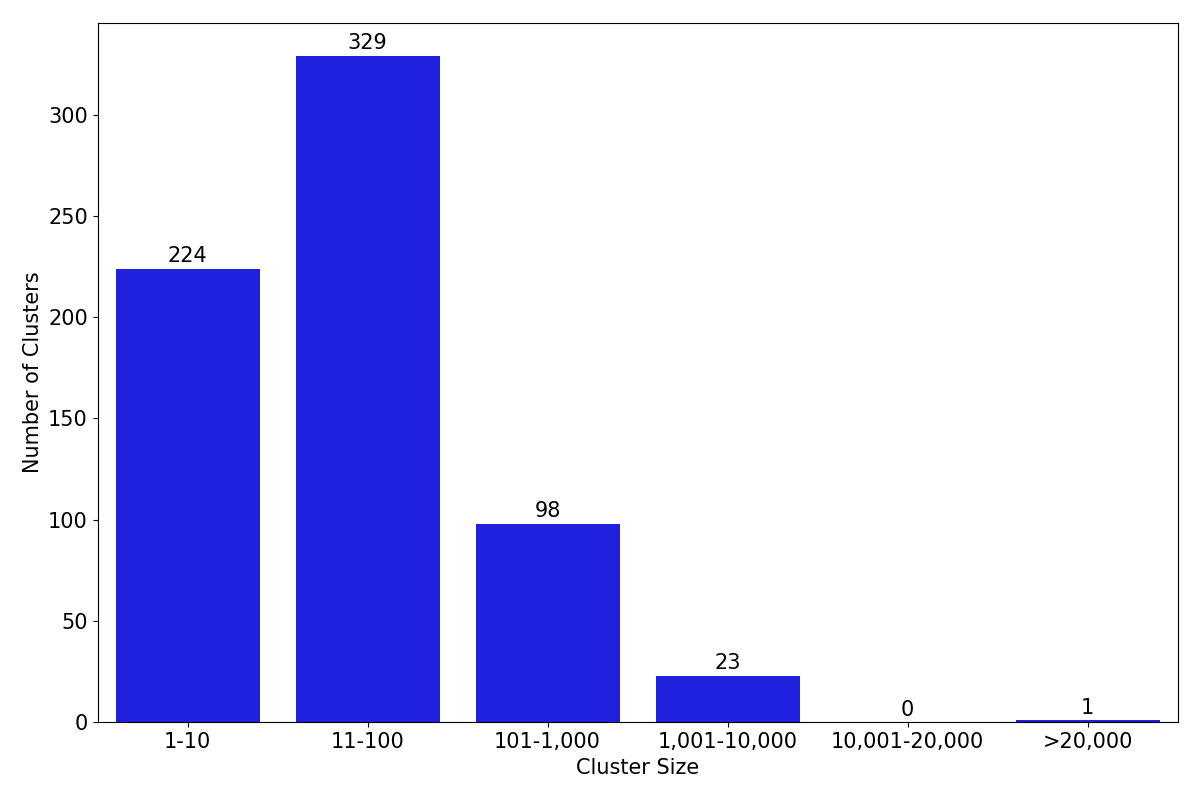}
    \caption{Histogram showing the number of clusters of images whose size is within each band. There is only one cluster with more than 20k images, this is a cluster of outliers, and it contains 60k images. 
    }
    \label{fig:clusters_size_bands}
    \centering
\end{figure}

We leverage the cluster structure used to remove duplicate posts to explore and overview images and topics in our dataset regarding the data analysis. Figure~\ref{fig:clusters_size_bands} shows a histogram with the number of clusters of images by size. In all, we have 675 clusters, where one cluster contains outliers. Then, to ease the visualization, we grouped the clusters by size, after removing duplications, considering the size ranges: $[1-10]$, $[11-100]$, $[101-1000]$, $[1001 - 10,000]$, $[10,001 - 20,000]$ and $20,001$ or more.

\begin{figure}[tp!]
    \includegraphics[width=0.6\textwidth]{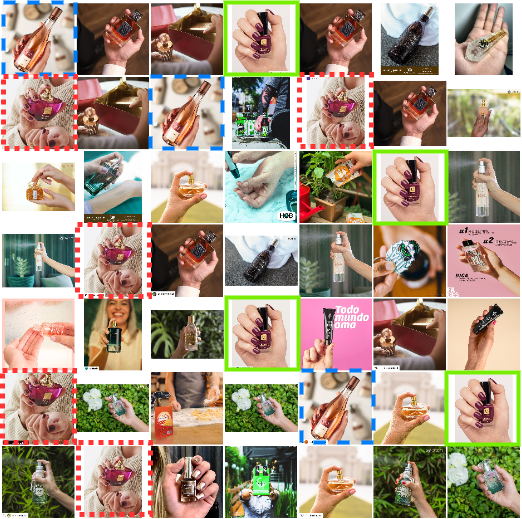}
    \caption{A sample of images from a cluster whose majority of the images are related to Perfumes. We highlighted the duplicated images, such that the ones with borders in the same color and line are considered duplications. Observe that there are many duplications in this cluster, and it is worth noting the importance of preprocessing these cases.}
    \label{fig:cluster_perfum}
    \centering
\end{figure}

Also, we draw a sample of images from each cluster to visualize and possibly create classes of images afterward. Here we present image samples from representative and well-formed clusters. Note that we drew these samples before removing duplications. For example, figure~\ref {fig:cluster_perfum} shows a cluster with images of many types of perfume and beauty products. We highlighted few duplicated images, such that the images with borders in the same color and line are considered duplications. Duplications are frequent in the clusters, and, as shown, one image may be duplicated more than once, and it is worth noting the importance of preprocessing these cases.

\begin{figure}[htp!]
    \includegraphics[width=0.6\textwidth]{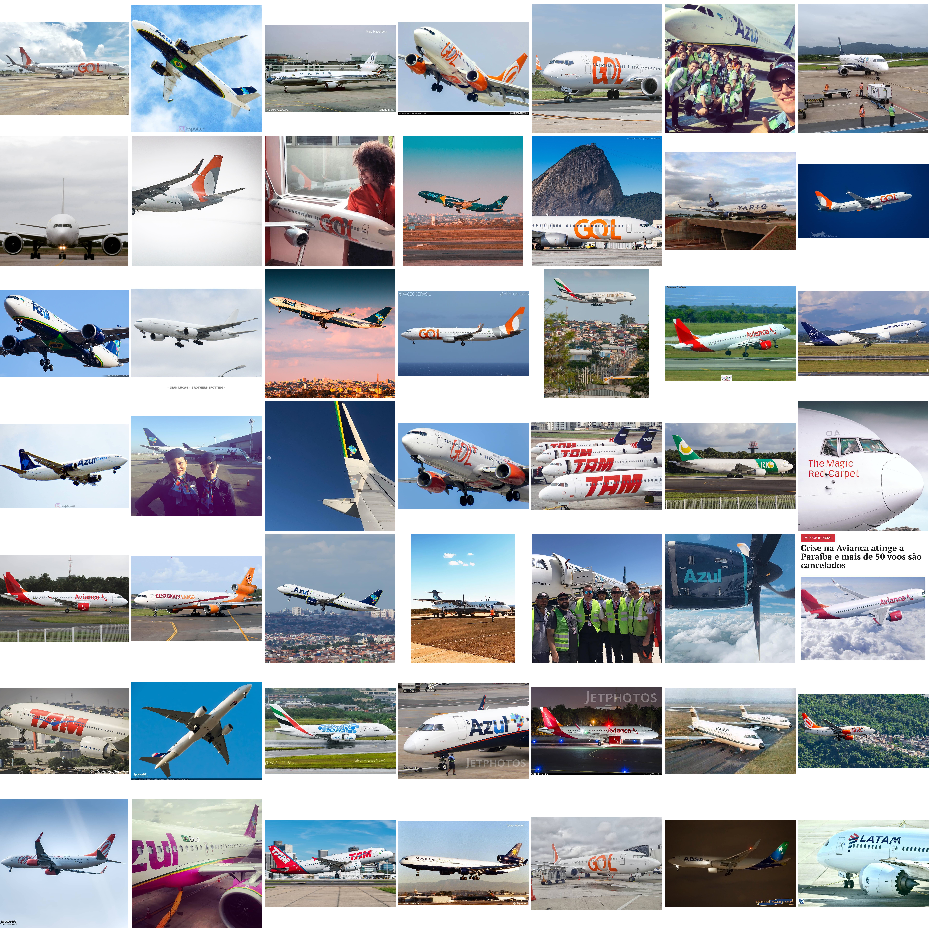}
    \caption{Sample from a cluster of airplanes. It is worth noting the variety of positions of the airplanes, some images show just part of them, such as the wings, the turbines, \textit{etc}.}
    \label{fig:cluster_airplanes}
    \centering
\end{figure}

The sample of Figure~\ref{fig:cluster_airplanes} shows images of airplanes in several places, with or without other surrounding elements, flying or not, \textit{etc}. It is worth noting the variety of aircraft positions, such as showing just the cockpit, wings, and turbine engines. This cluster is well-formed and consists of a representative set of images of airplanes.

\begin{figure}[htp!]
    \includegraphics[width=0.6\textwidth]{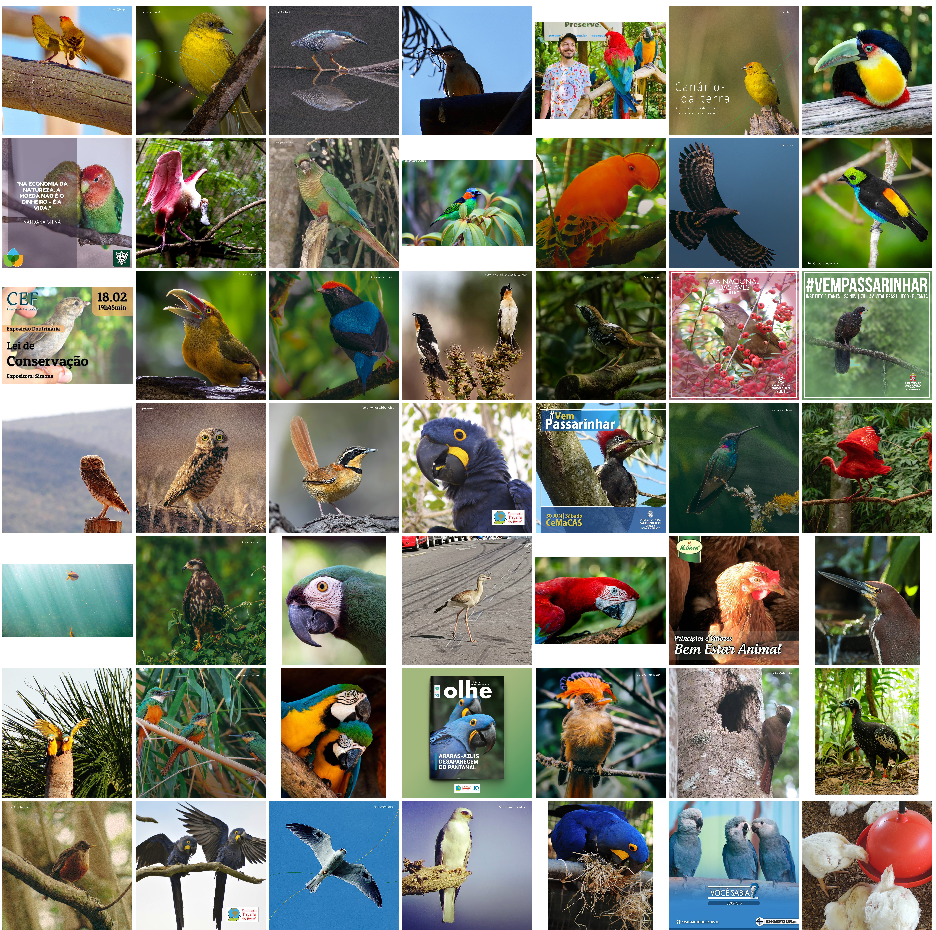}
    \caption{Samples from a cluster of birds. There is a diversity of species of birds as well as a variety of number of animals in the photographs.}
    \label{fig:cluster_birds}
    \centering
\end{figure}

Figure~\ref{fig:cluster_birds} shows a cluster that depicts birds from different species, in a diversity of positions, flying or not. Although a few images in this cluster illustrate advertisement campaigns, as can be seen by the texts in them, the images have in common the fact that they show the birds as the primary elements.

\begin{figure}[htp!]
    \includegraphics[width=0.6\textwidth]{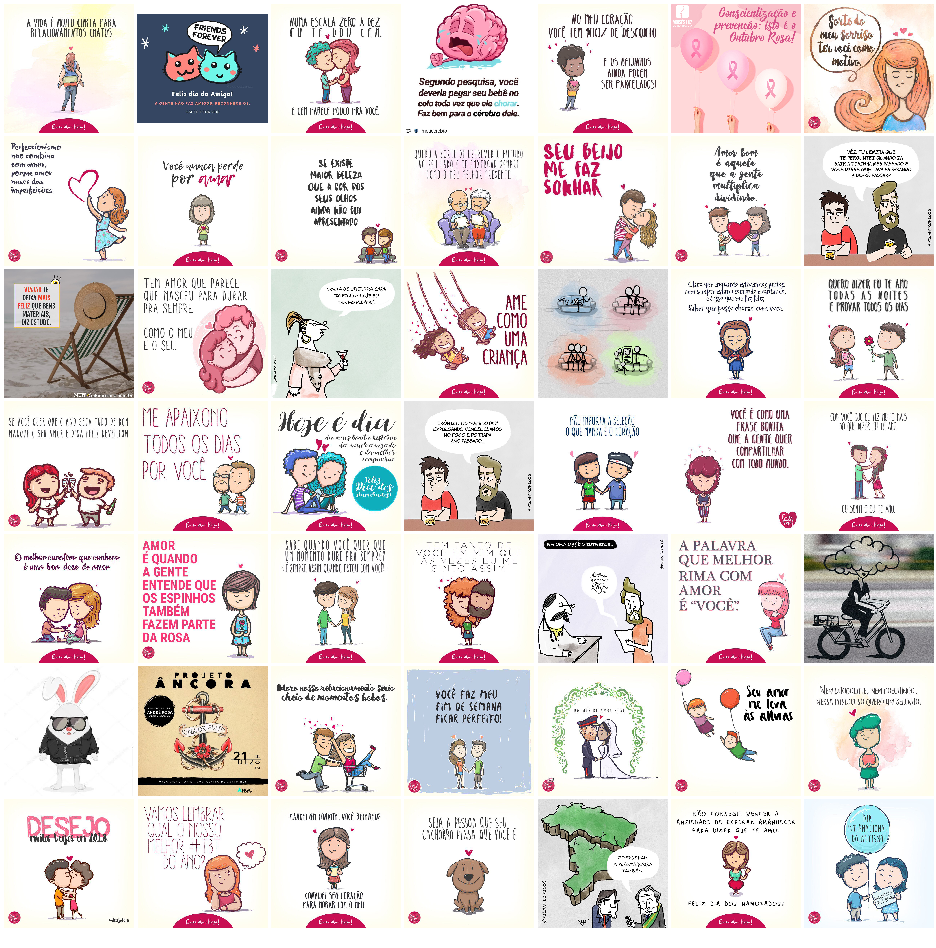}
    \caption{Samples from a cluster of cartoons. It worth noting that most of the cartoons are made by the same author, and thus they have the same stroke.}
    \label{fig:cluster_cartoon1}
    \centering
\end{figure}

\begin{figure}[htp!]
    \includegraphics[width=0.6\textwidth]{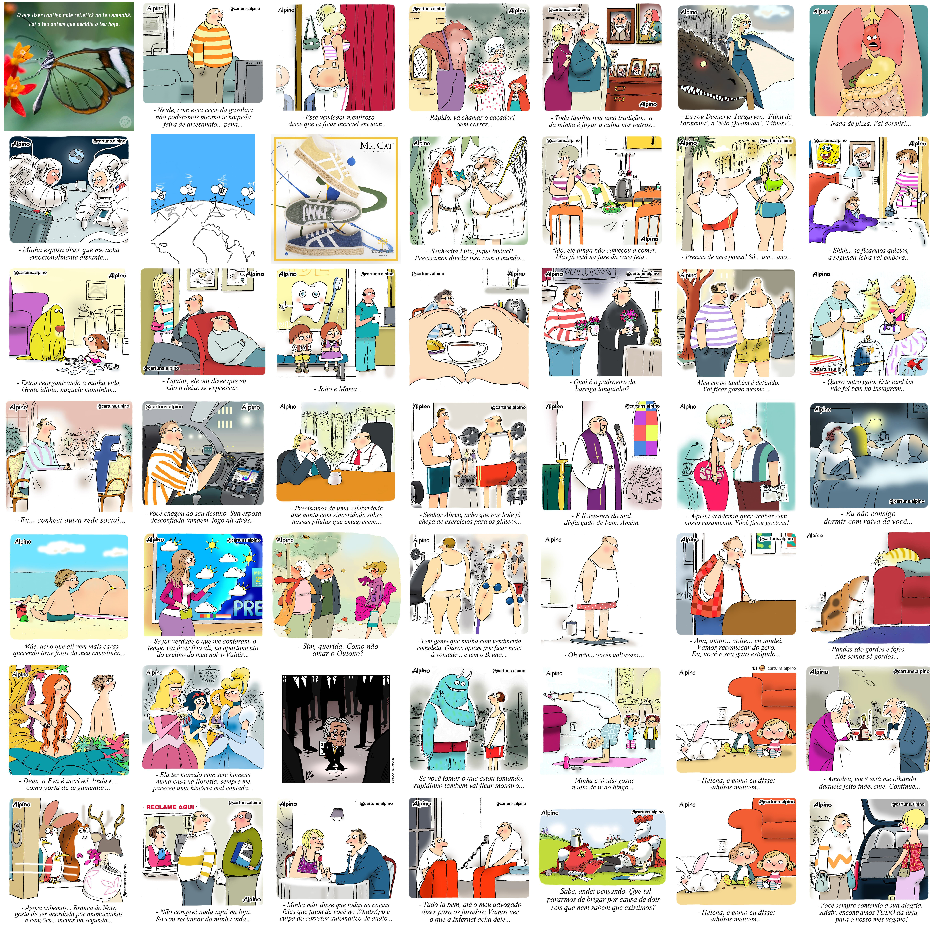}
    \caption{Samples from a cluster of cartoons. Note that despite the images present in this cluster being cartoons, they have a different style from the ones in Figure~\ref{fig:cluster_cartoon1}.}
    \label{fig:cluster_cartoon2}
    \centering
\end{figure}

As we can see both clusters illustrated by Figures \ref{fig:cluster_cartoon1} and \ref{fig:cluster_cartoon2} include images of cartoons. Note that these figures represent samples from different clusters. However, it may be observed that the cartoons of Figure~\ref{fig:cluster_cartoon1} are different from the ones in Figure~\ref{fig:cluster_cartoon2}. We can note differences in stroke, colors, and style.

\begin{figure}[htp!]
    \includegraphics[width=0.6\textwidth]{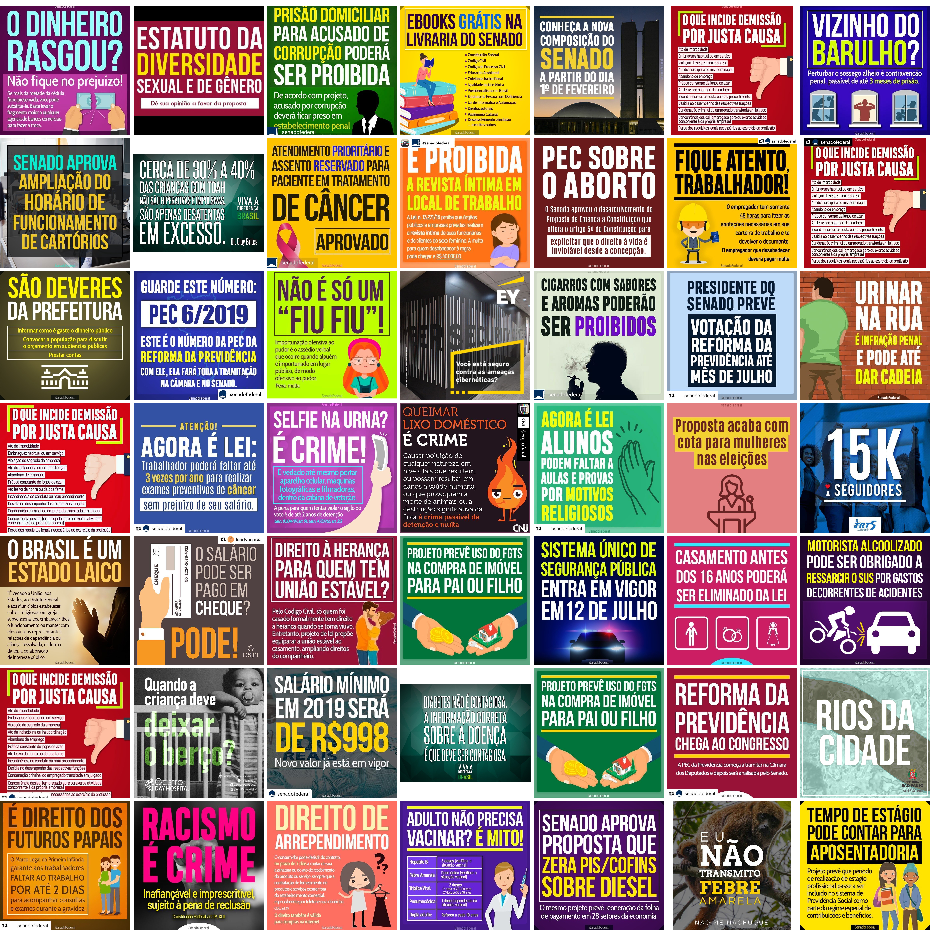}
    \caption{Samples from a cluster of informative texts. This cluster illustrates draft laws presented in Brazilian Congress.}
    \label{fig:cluster_ads}
    \centering
\end{figure}

The cluster shown in Figure~\ref{fig:cluster_ads} consists of advertisements that illustrate draft laws presented in the Brazilian Congress. This type of post accounts for a significant part of this dataset because many government agencies and official politicians' profiles frequently publish posts in this format.

\begin{figure}[htp]
    \centering
    \subfloat[Topic of captions related to Beach.]{{\includegraphics[scale = 0.33]{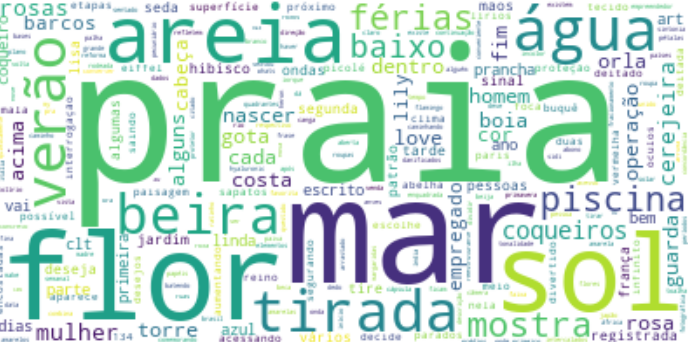} }}\hspace{0.01cm}
    \subfloat[Topic of captions related to Family.]{{\includegraphics[scale = 0.33]{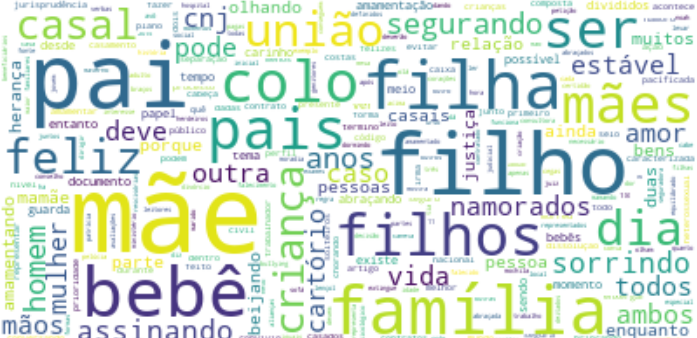} }}
    \qquad
    \subfloat[Topic of captions related to Fake News.]{{\includegraphics[scale = 0.33]{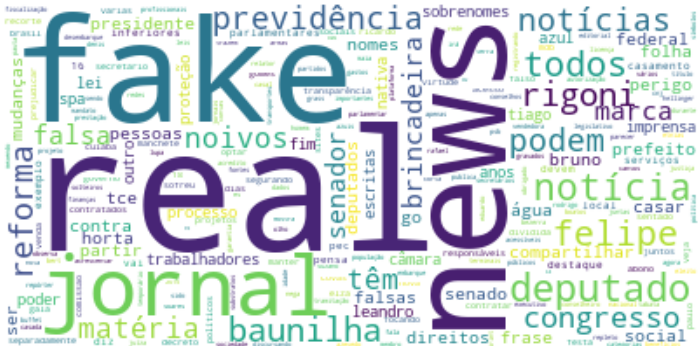} }}\hspace{0.01cm}
    \subfloat[Topic of captions related to Electoral Justice.]{{\includegraphics[scale = 0.33]{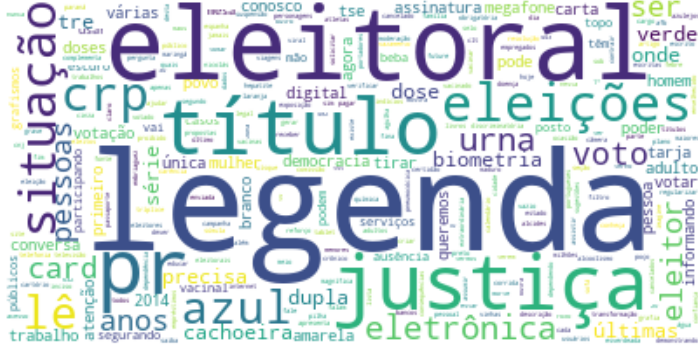} }}
    \qquad
    \subfloat[Topic of captions related to Disabled People.]{{\includegraphics[scale = 0.33]{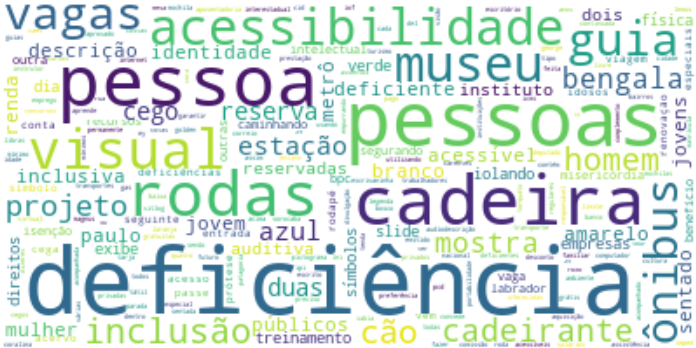} }}\hspace{0.01cm}
    \subfloat[Topic of captions related to Cosmetic Products.]{{\includegraphics[scale = 0.33]{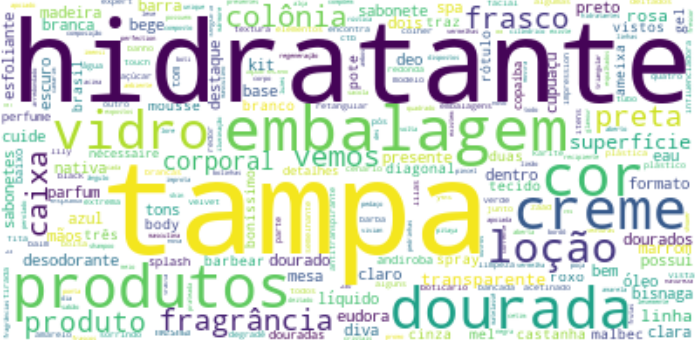} }}
    
    \caption{Word clouds showing the most frequent words in each topic found in the dataset. The topics were modeled using the LDA algorithm. It can be identifiable topics related to Beach, Family, Fake News, Electoral Justice, Disabled People, and Cosmetic Products.}
    \label{fig:topics}
\end{figure}

Moreover, to visualize the general topics occurring in captions of \textit{\#PraCegoVer} dataset, we carried statistical modeling, Topic Modeling, using Latent Dirichlet Allocation (LDA) algorithm \cite{lda2003}. This kind of analysis allows us to understand our data better. In Figure~\ref{fig:topics} we present word clouds of the most interpretable topics. Also, in these word clouds, we show the most frequent words for each~topic. 

In Figure \ref{fig:topics} we illustrate few topics in our dataset. Topic (a) shows elements of beach such as sand, sea, sun, and beach. 
We can see that Topic (b) is about Family because it comprehends words such as father, mother, baby, children, \textit{etc}.
In Topic (c) the words ``fake'' and ``news'' are the most often followed by ``real'' and ``newspaper'', indicating it is related to ``Fake News''. 
Topic (d) illustrates words related to Justice, specifically the Brazilian Electoral Justice, as can be seen in the terms ``Electoral'', ``Elections'', ``Justice'', and ``Allegiances''.
Regarding Topic (e), it is based on concepts related to Disability, such as wheelchair, accessibility, and inclusion.
Topic (f) is related to Cosmetic Products, having frequent words such as moisturizing cream, packaging, fragrance, perfume, \textit{etc}.

\subsection{Comparative Analysis}
\label{sub:comparative_analysis}
This section describes our dataset's statistics compared to those of the MS COCO dataset since it is by far the most used dataset.

\begin{figure}[tp]
    \includegraphics[width=0.5\textwidth]{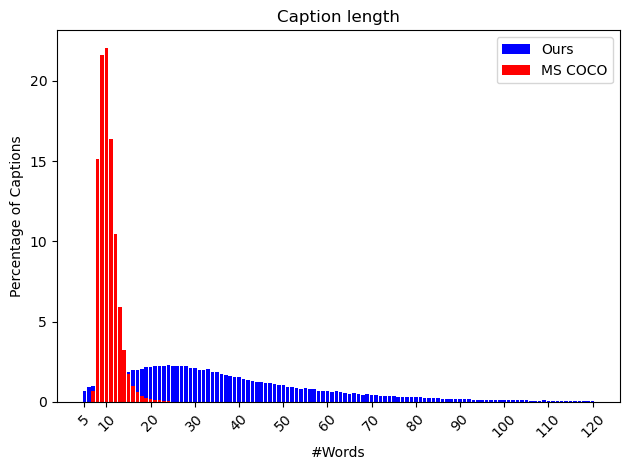}
    \caption{Histogram of the distribution of captions by length in terms of number of words. We plot the caption length distribution for \textit{\#PraCegoVer} (blue) and MS COCO (red) datasets.}
    \label{fig:histogram_description_ditribuition}
    \centering
\end{figure}

Figure~\ref{fig:histogram_description_ditribuition} shows the distribution of descriptions by length, in terms of number of words, in \textit{\#PraCegoVer} and MS COCO datasets. The descriptions in our dataset have, on average, roughly 40 words, whereas MS COCO has only about~10. Also, the caption length variance in \textit{\#PraCegoVer} is greater than in MS COCO. These two characteristics make our dataset more challenging than MS COCO.
Still, considering that the most employed evaluation metric in Image Captioning Literature,  CIDEr-D \cite{CIDEr2015}, relies on MS COCO to set its hyperparameters, it does not work well in datasets where the caption length differ significantly from MS COCO. Moreover, it is essential to highlight that the majority of state-of-the-art models are trained using Self-Critical Sequence Training~\cite{rennie2017self}, a Reinforcement Learning approach that aims to maximize CIDEr-D. Thus, the mean and variance of caption length in a dataset play an essential role in the final result.

\begin{figure}[tp]
    \includegraphics[width=0.5\textwidth]{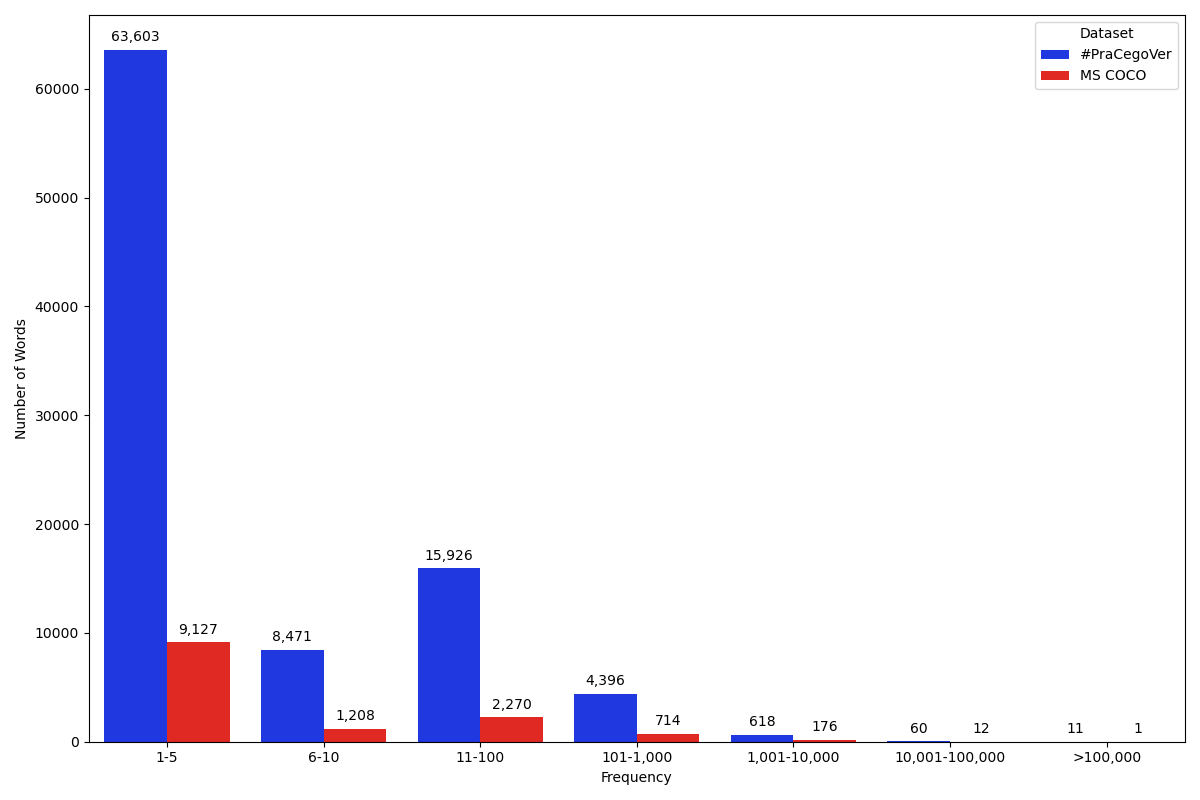}
    \caption{Histogram of word frequency of \textit{\#PraCegoVer} (blue) and MS COCO (red) datasets. We plot the number of words for each considering frequency range.}
    \label{fig:word_freq_mscoco_pracegover}
    \centering
\end{figure}

Moreover, we plot in Figure \ref{fig:word_freq_mscoco_pracegover} the distribution of words by frequency, \textit{i.e.}, the number of occurrences of that word, for \textit{\#PraCegoVer} and MS COCO datasets. On the x-axis, we show ranges of word frequency, and on the y-axis, we show the number of words whose frequency is within that band. As we can note, our dataset has by far more words occurring five or fewer times in the captions. Considering words as classes predicted by the models, if we train such models, they will ``see'' only a few examples of words with low frequency, then the models will not learn those classes. Therefore, this characteristic also makes our dataset more challenging than MS COCO.

\section{Experiments}
\label{sec:experiments}

To validate our dataset and have a benchmark, we carried out experiments with AoANet \cite{AoANet_2019}, one of the state-of-art algorithms for MS COCO Captions. AoANet is based on the Attention on Attention (AoA) module to determine the importance of attention results given queries. AoA module introduces an ``Attention Gate'' and an ``Information Vector'' to address the attention mechanism's issue, returning irrelevant results to the query. In our experiments,
we selected a subset with 100,000 of our collected posts that, after being cleaned, results in 62,935 pairs of image-caption (see Section \ref{sec:preprocessing}), we called it as \textit{\#PraCegoVer}-100k. Then, we split the set into train, validation and test sets (see Section \ref{sub:dataset splits}) whose are 37,881, 12,442, 12,612, respectively. We trained and validated AoANet models on MS COCO Captions and \textit{\#PraCegoVer}-100k, considering the same hyperparameters. We trained the models firstly to optimize the Cross-Entropy Loss and then directly maximizing CIDEr-D score  \cite{CIDEr2015} using Self-Critical Sequence Training (SCST) \cite{rennie2017self}.
We evaluated the models considering the same metrics used on MS COCO competition: BLEU \cite{BLEU2002}, ROUGE \cite{ROUGE2004}, METEOR \cite{METEOR2005} and CIDEr-D \cite{CIDEr2015}.

\subsection{Results and Analysis}

Table \ref{table:results} shows that the performance of the model trained on MS COCO Captions drops considerably for all evaluation metrics compared to the one trained on \textit{\#PraCegoVer}-100k. We expected this result because of the difference in mean and variance of caption length between our dataset and MS COCO, as we discussed in Section~\ref{sub:comparative_analysis}. Using the SCST approach to maximize the CIDEr-D score makes the models learn more easily sentences with similar size because CIDEr-D takes sentence size directly into account, and predicted sentences that differ in length from the reference are hardly penalized. Moreover, the longer the sentences a model predicts, the more likely they contain words that are not in the reference. Thus, the models trained with SCST combined with the CIDEr-D score learn to predict short sentences. However, as our reference descriptions have 40 words on average, the performance is expected to be~poor. 

\begin{table}[htp!]
\caption{Experimental results obtained by training AoANet model on \textit{\#PraCegoVer}-100k and MS COCO Captions. The performance of the model trained on MS COCO Captions drops considerably for all evaluation metrics compared to the one trained on \textit{\#PraCegoVer}-100k. This result is expected because of the difference in mean and variance of caption length between our dataset and MS COCO, as we discussed in Section~\ref{sub:comparative_analysis}.}
\label{table:results}
\begin{tabular}{lcccc}
\toprule
\multicolumn{1}{c}{{\color[HTML]{000000} \textbf{Dataset}}} &
  {\color[HTML]{000000} \textbf{CIDEr-D}} &
  {\color[HTML]{000000} \textbf{ROUGE-L}} &
  {\color[HTML]{000000} \textbf{METEOR}} &
  {\color[HTML]{000000} \textbf{BLUE-4}} \\ \toprule
{\color[HTML]{000000} MS COCO Captions} &
  {\color[HTML]{000000} 120.8} &
  {\color[HTML]{000000} 57.6} &
  {\color[HTML]{000000} 27.8} &
  {\color[HTML]{000000} 36.7} \\ 
{\color[HTML]{000000} \textit{\#PraCegoVer}-100k} &
  {\color[HTML]{000000} 3.2} &
  {\color[HTML]{000000} 13.4} &
  {\color[HTML]{000000} 6.5} &
  {\color[HTML]{000000} 1.6} \\ \bottomrule
\end{tabular}
\end{table}

\subsection{Qualitative Analysis}

\begin{figure}[htp!]
\centering
    \subfloat[
    \textbf{Reference}:  Na foto, Thalita Gelenske e Thaís Silva estão abraçadas com Luana Génot na livraria Travessa. Ao fundo, diversos livros coloridos estão na prateleira. Nas laterais da foto, existem 2 banners: um deles vermelho, com o logo da, e o outro com a divulgação do livro da Luana.''; 
    \textbf{Generated}: ``Foto de uma mulher segurando um livro com livros. Ao fundo, uma estante com livros. Texto: “A sua. É sua festa. É sua!”.'';]{\includegraphics[width=0.23\textwidth]{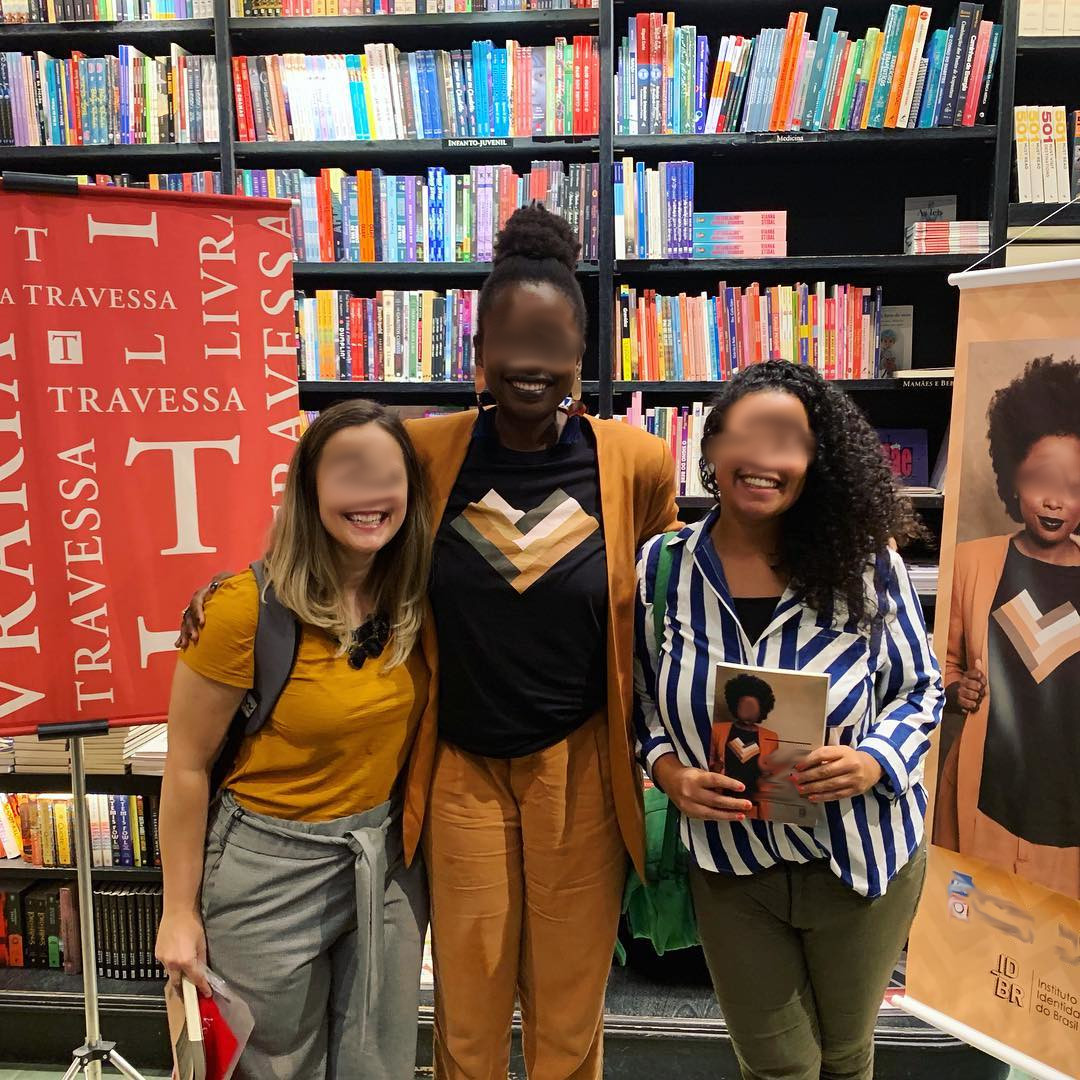}\label{fig:generated_descriptions_library}}\hspace{0.2cm}
    \subfloat[\textbf{Reference}: ``Em um ambiente externo, uma gata de pelagem branca e caramelo, está deitada de olhos fechados, sua ninhada de filhotinhos esta ao seu redor, um dos gatinhos esta olhando para a câmera, ele tem olhos cor de mel e pelagem branca, preta e caramelo, o restante dos filhotes, estão desfocados. No canto inferior direito, está escrito, “PremieRpet” em letras alaranjadas.''; 
    \textbf{Generated}: ``Foto de um gatinho de gato gato em uma gatinha de pelagem.';]{\includegraphics[width=0.23\textwidth]{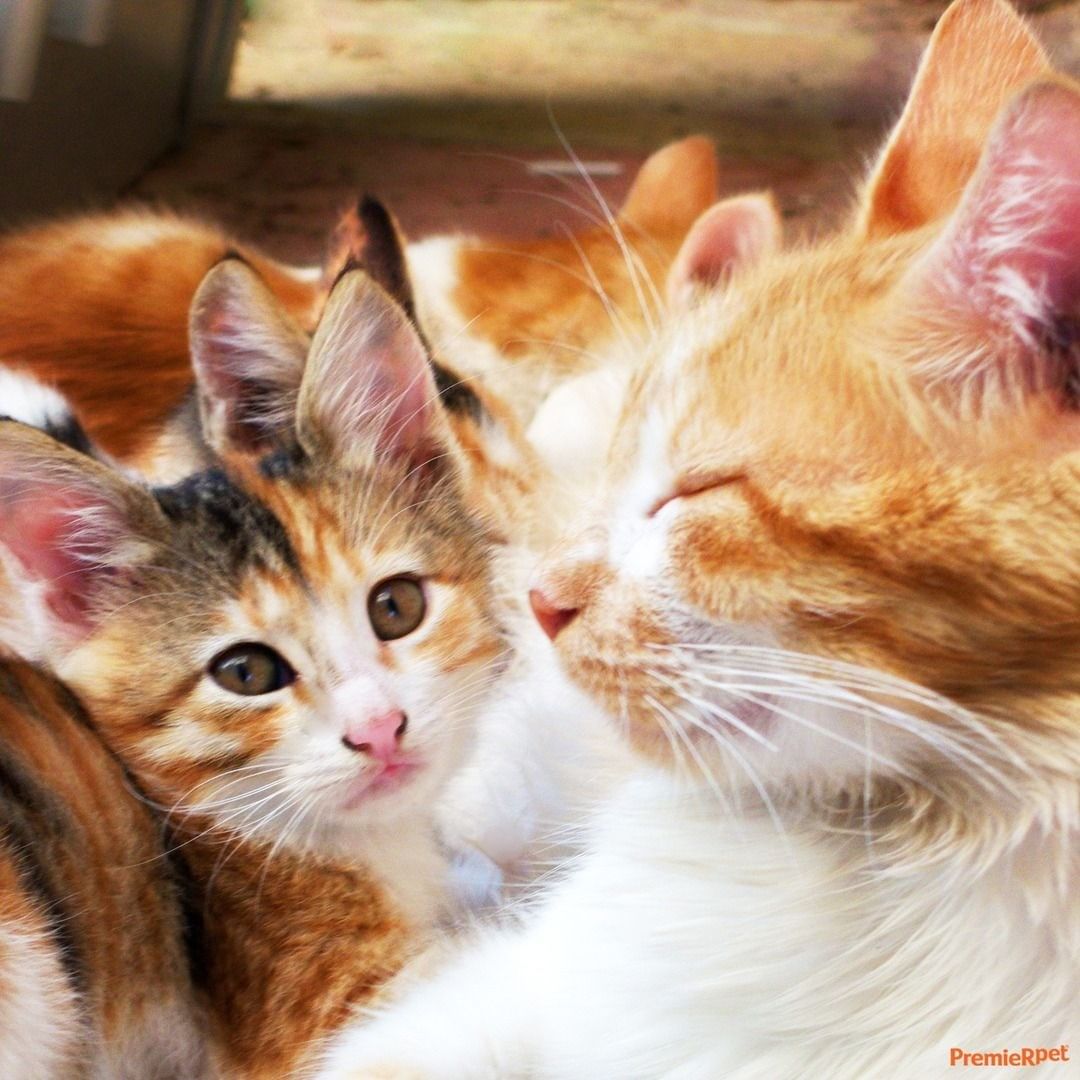}\label{fig:generated_descriptions_cat}}\hspace{0.2cm}
    \subfloat[\textbf{Reference}: ``Fotografia aérea sobre o pedágio da Terceira Ponte. A foto contém alguns prédios, um pedaço da Terceira Ponte e o fluxo de carros.''; 
    \textbf{Generated}: ``foto aérea aérea aérea da cidade de Florianópolis mostrando casas casas, mostrando casas casas. Ao fundo, algumas casas e casas.';]{\includegraphics[width=0.23\textwidth]{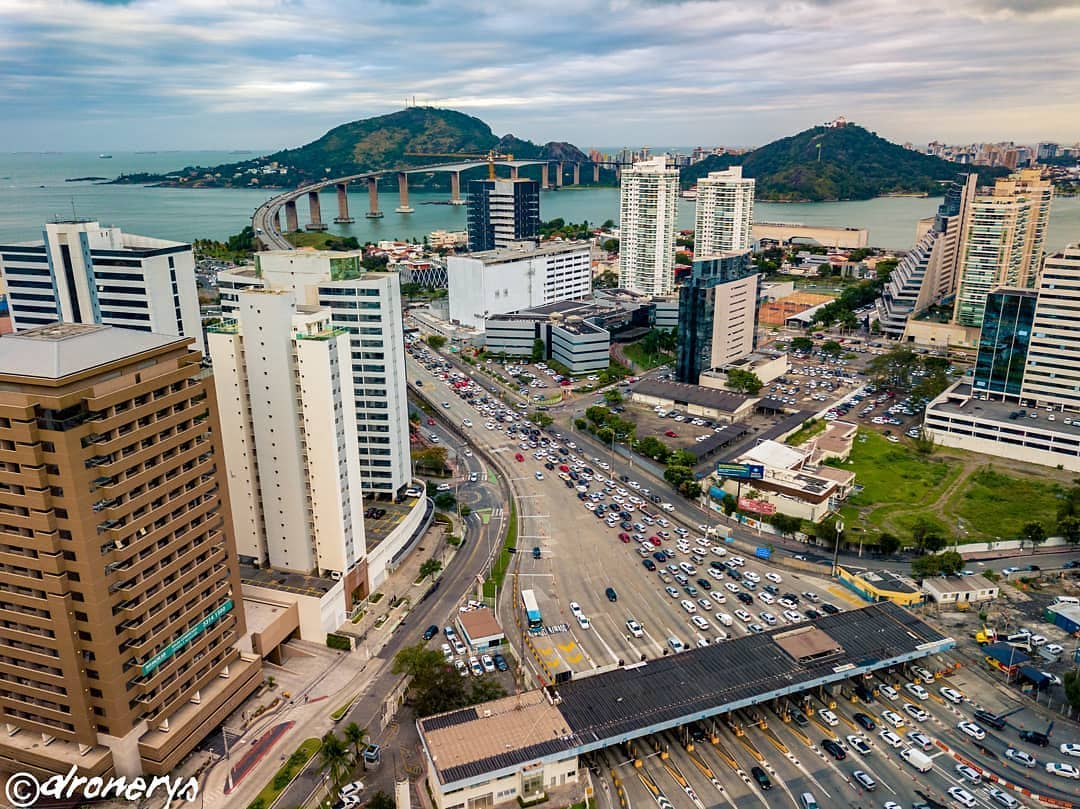}\label{fig:generated_descriptions_city}}\hspace{0.2cm}
    \subfloat[\textbf{Reference}: ``Quadrado laranja. Ao centro o texto em cor branca: 13ª Semana pela paz em casa. Em ambos os lados pequenas barras na cor azul com os respectivos dados: 2.333 processos movimentados. 610 sentenças. 552 despachos. 348 medidas protetivas. 239 audiências.''; 
    \textbf{Generated}: ``Imagem com fundo amarelo. Texto: “: você tem o que você ?”. Texto: “: você tem direito: você tem direito: você tem direito: você tem direito: R \$ 10 \% ;: “;: “;: “;: “;: “; ; ; ; ; ; ;;';]{\includegraphics[width=0.23\textwidth]{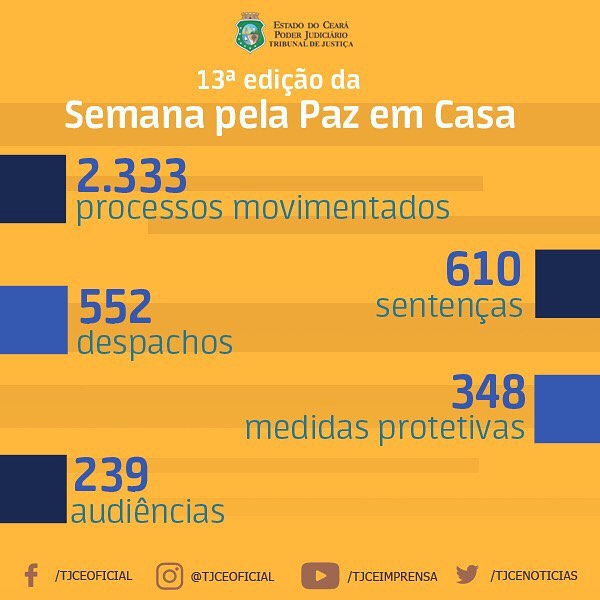}\label{fig:generated_descriptions_ads}}
    
    \caption{Examples of images followed by their reference captions and the descriptions generated by the model trained on subset \textit{\#PraCegoVer}-100k optimizing CIDEr-D.}
    \label{fig:generated_descriptions}
\end{figure}

Figure~\ref{fig:generated_descriptions} illustrates some images of \textit{\#PraCegoVer}-100k with their reference captions and the descriptions generated by the model trained on our dataset. It is worth noting that the model predicted incomplete sentences most of the time, as illustrated in Figure~\ref{fig:generated_descriptions_library}. Also, it generates sentences where the same word is repeated many times, as can be seen clearly in Figures~\ref{fig:generated_descriptions_cat} and \ref{fig:generated_descriptions_city}. Finally, regarding advertisements where textual information is often present in the images, the model can not capture the texts on them and predicts meaningless punctuation signs, as shown in Figure~\ref{fig:generated_descriptions_ads}.

\section{Conclusions and Future Work}
\label{sub:conclusion}

Describing image content using natural sentences is an important task to make the Internet more inclusive and democratic. There are many datasets for the problem of image captioning, but the majority only consider English captions. Thus, this paper introduced the \textit{\#PraCegoVer} dataset, the first dataset for image captioning in Portuguese. Our dataset is based on audio descriptions created by supporters of the PraCegoVer initiative. The data comprehends various images and topics, representing real-world images posted on social media. It has characteristics that make it more challenging than MS COCO Captions, such as long sentences and many words occurring few times in the captions. 

In this paper, we compared AoANet models trained on MS COCO Captions and \textit{\#PraCegoVer}-100k, and we showed that it had poor results in our dataset due to the SCST approach. Furthermore, we developed a framework for post-collection from a hashtag on Instagram. Also, we proposed an algorithm to cluster and remove post duplication based on visual and textual information. Thus, we aim to contribute to Portuguese speakers' vision community and inspire other works on image captioning for non-English descriptions.

For future work, we intend to explore the influence of the number of words per image and the word frequency in models' learning and generalization capacity. Besides, we plan to remove proper names from the descriptions and analyze the impacts. Finally, we will collect more data to release different versions of our dataset as it grows over time. 


\section*{Acknowledgements}

G. O. dos Santos is funded by the São Paulo Research Foundation (FAPESP) (2019/24041-4)\footnote{The opinions expressed in this work do not necessarily reflect those of the funding agencies.}. 
S. Avila is partially funded by~FAPESP (2013/08293-7), a CNPq PQ-2 grant (315231/2020-3), and Google LARA~2020. 


\bibliographystyle{acm}
\bibliography{references}

\appendix
\newpage
\renewcommand\thefigure{\thesection.\arabic{figure}}    
\section{\emph{\#PraCegoVer} dataset}\label{app:datasheets}
Here, we present a detailed description of the dataset, a datasheet for the \emph{\#PraCegoVer} dataset, as proposed by Gebru et al.~\cite{gebru2020datasheets}.

\subsection*{Motivation}

\subsubsection*{For what purpose was the dataset created? Was there a specific task in mind?}
\emph{\#PraCegoVer} dataset has been created to provide images annotated with descriptions in Portuguese for the image captioning task. With this dataset, we aim to alleviate the lack of datasets with Portuguese captions for this task.
 
\subsubsection*{Who created the dataset?}
\emph{\#PraCegoVer} dataset was created by G.O.S., E.L.C., and S.A., on behalf of the Institute of Computing at the University of Campinas (Unicamp).

\subsubsection*{Who funded the creation of the dataset?}
The creation of \emph{\#PraCegoVer} dataset is partially funded by FAPESP grant (2019/24041-4).

\subsection*{Composition}

\subsubsection*{What do the instances that comprise the dataset represent?}
The instances represent public posts collected from Instagram tagged with \#PraCegoVer, comprising images and captions.

\subsubsection*{How many instances are there in total?}
The dataset comprehends 520,997 instances.

\subsubsection*{What data does each instance consist of?}
Each instance consists of the post and collection dates, an identifier for the post owner (anonymized), an image, a raw caption (as originally written by the post author), and an audio description extracted from the raw caption that describes the image content.

\subsubsection*{Is there a label or target associated with each instance?}
The image captioning task consists of generating captions for images, thus the label of each instance is represented by its caption (\textit{i.e.}, audio description).

\subsubsection*{Are there recommended data splits?}
The dataset comes with two specified train/validation/test splits, one for \textit{\#PraCegoVer}-63k (train/validation/test: 37,881/12,442/ 12,612) and another for \textit{\#PraCegoVer}-173k (train/validation/test:  104,004/34,452/34,882). These splits are subsets of the whole dataset.


\subsubsection*{Are there any errors, sources of noise, or redundancies in the dataset?}
\textit{\#PraCegoVer} dataset relies on data labeled in the wild, and audio descriptions are automatically extracted from raw captions using regular expressions. Thus, audio descriptions are susceptible to errors that are inherent to the source. Still, many posts are re-posted on Instagram, changing just a few details in the image and text. Then, some instances are very similar, and we annotate them as duplications.

\subsubsection*{Does the dataset contain data that might be considered confidential?}
No, we only collect data as public by their owners.

\subsubsection*{Does the dataset contain data that, if viewed directly, might be offensive, insulting, threatening, or might otherwise cause anxiety?}
An initial analysis shows that although there exist words that can be offensive, as illustrated in Table \ref{table:offensive_words}, they are insignificant because they occur rarely. Note that overall such words occupy the position 30,000th or lower in the rank of word frequency. The dataset consists of data collected from public profiles on Instagram that were not thoroughly validated. Thus there might be more examples of offensive and insulting content.

\subsubsection*{Is it possible to identify individuals either directly or indirectly?}
The dataset consists of data collected from public profiles on Instagram. There are many examples of images of people, and thus the individuals present in those images can be easily identified.

\subsubsection*{Does the dataset contain data that might be considered sensitive in any way?}
The dataset consists of data collected from public profiles on Instagram. Therefore, the images and raw captions might contain data revealing racial or ethnic origins, sexual orientations, religious beliefs, political opinions, or union memberships.

\subsection*{Collection Process}

\subsubsection*{How was the data associated with each instance acquired?}
We collected the data directly from Instagram, and they were not entirely validated. Also, we extracted the audio description from the raw caption by using regular expressions.

\subsubsection*{What mechanisms or procedures were used to collect the data?}
We have implemented an automated crawler using Selenium\footnote{\url{https://selenium-python.readthedocs.io}} and Selenium Wire\footnote{\url{https://github.com/wkeeling/selenium-wire}} that scrolls over the profile page collecting the data from posts, whose caption is tagged with the \textit{\#PraCegoVer}. We execute this process daily and incrementally, storing the images, captions, post date, and the collection date. 

\subsubsection*{Who was involved in the data collection process and how were they compensated?}
The users from Instagram spontaneously published the posts, and we collected them using a crawler developed by the authors of this paper. Thus, regarding the collection process, the student, G.O.S., is compensated with the scholarship funded by the São Paulo Research Foundation (FAPESP) (2019/24041-4).

\subsubsection*{Over what timeframe was the data collected? Does this timeframe match the creation timeframe of the data associated with the instances?}
We daily collect the posts that were created any time ago. However, since we execute daily the collection process, most timeframes of the data match the creation timeframe.

\subsubsection*{Were any ethical review processes conducted?}
No ethical review processes were conducted.

\subsubsection*{Did you collect the data from the individuals in question directly, or obtain it via third parties or other sources?}
We collected data from Instagram using a crawler, thus via third parties.

\subsubsection*{Were the individuals in question notified about the data collection?}
Since the data were automatically collected from public profiles, the individuals were not notified.

\subsubsection*{Did the individuals in question consent to the collection and use of their data?}
The individuals that have public profiles on Instagram consent to the use of their data once they accept the Data Policy of the platform\footnote{\url{https://www.facebook.com/help/instagram/519522125107875/?helpref=hc\_fnav&bc[0]=Instagram\%20Help&bc[1]=Policies\%20and\%20Reporting}}, thus they consent to have their data accessed and downloaded through third-party services. Notwithstanding, we have not notified the individuals.

\subsubsection*{Has an analysis of the potential impact of the dataset and its use on data subjects been conducted?}
Such analysis was not conducted. However, we conducted an initial analysis of the bias within our dataset. Please, refer to \ref{app:uses} (Question: \textit{Is there anything about the composition of the dataset or the way it was collected and preprocessed that might impact future uses?})
 
\subsection*{Preprocessing}
 
\subsubsection*{Was any preprocessing of the data done?}
We preprocess the raw caption by using regular expressions to extract the audio description part within the caption. Also, we use the Duplication Clustering Algorithm (see Algorithm \ref{algo:grouping_dup}) to cluster the instances of duplicated posts, but instead of removing the duplications, we only annotate them.

\subsubsection*{Was the “raw” data saved in addition to the preprocessed data?}
In addition to the preprocessed data, we provide all the raw data.

\subsubsection*{Is the software used to preprocess the instances available?}
All the scripts used to preprocess the data are available on the repository of this project (\url{https://github.com/larocs/PraCegoVer}).

\subsection*{Uses}\label{app:uses}

\subsubsection*{Has the dataset been used for any tasks already?}
This dataset has been used for the image captioning task.

\subsubsection*{Is there a repository that links to any or all papers or systems that use the dataset?}
No.

\subsubsection*{What (other) tasks could the dataset be used for?}
This dataset could be used for image classification, text-to-image Generation, and sentiment analysis of the posts concerning specific periods such as electoral periods.

\subsubsection*{Is there anything about the composition of the dataset or the way it was collected and preprocessed that might impact future uses?}

\begin{figure}[h]
    \centering
    \subfloat[Word cloud related to women.]{{\includegraphics[scale = 0.2]{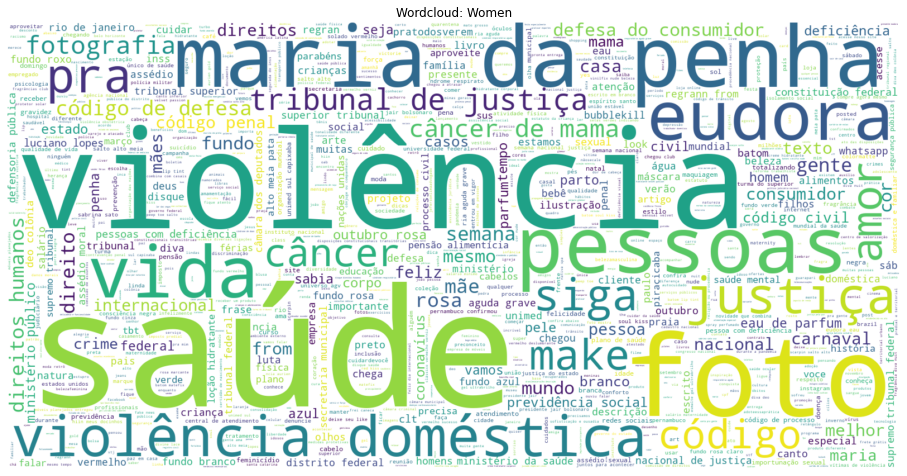} }}\hspace{0.05cm}
    \subfloat[Word cloud related to men.]{{\includegraphics[scale = 0.2]{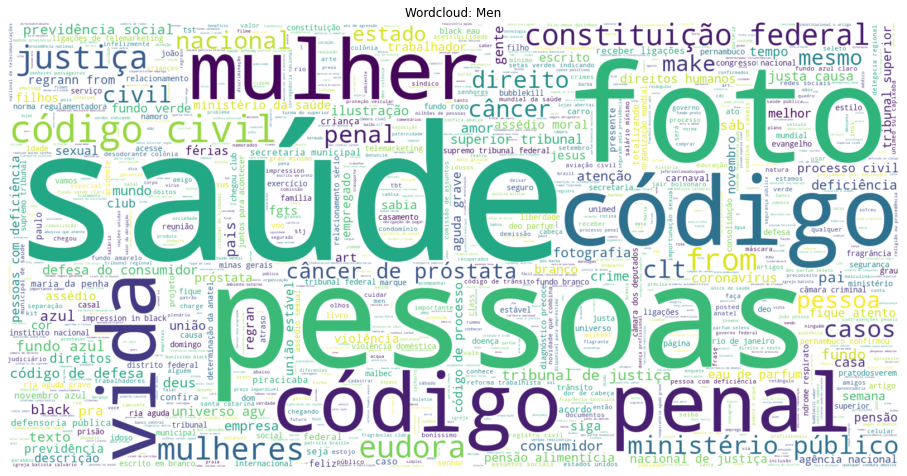} }}
    
    \subfloat[Word cloud related to black women.]{{\includegraphics[scale = 0.2]{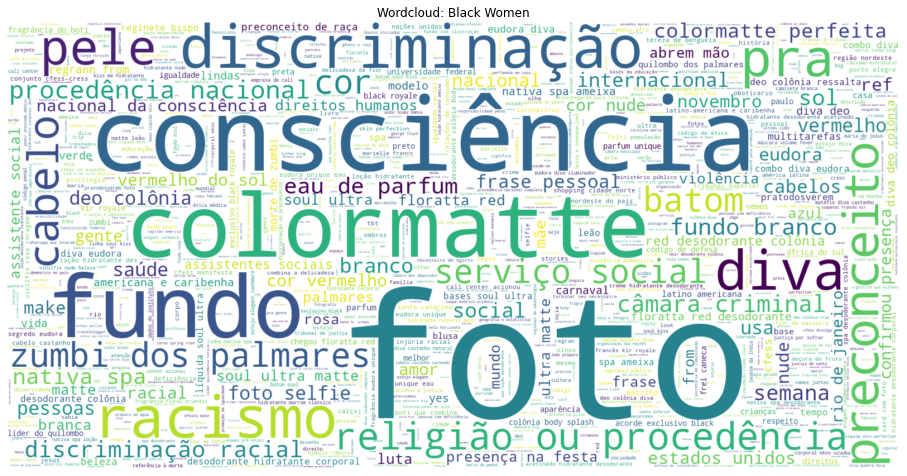} }}\hspace{0.05cm}
    \subfloat[Word cloud related to black men.]{{\includegraphics[scale = 0.2]{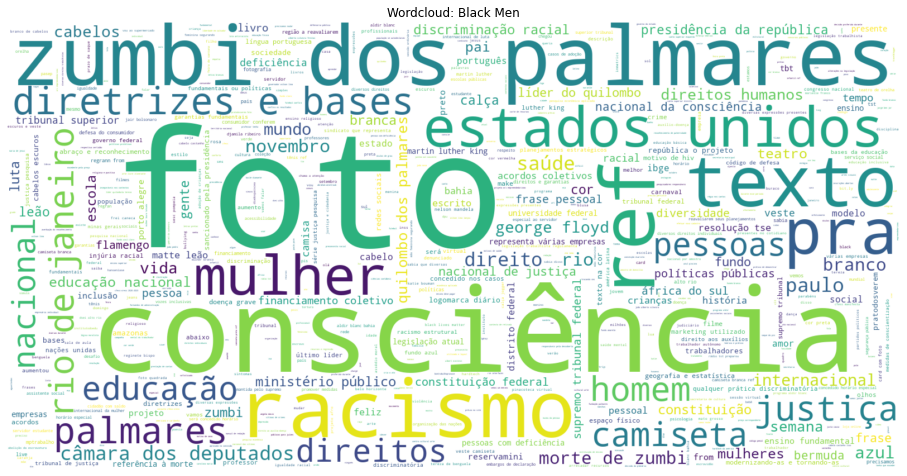} }}
   
    \subfloat[Word cloud related to indigenous women.]{{\includegraphics[scale = 0.2]{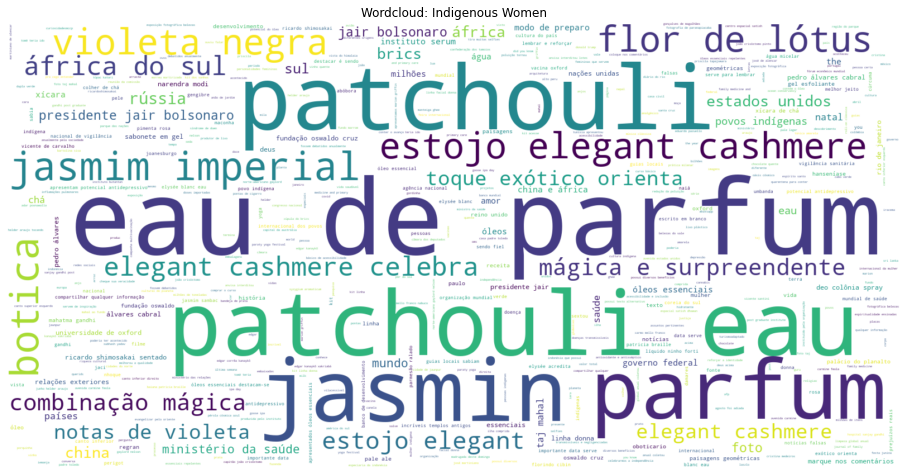} }}\hspace{0.05cm}
    \subfloat[Word cloud related to indigenous men.]{{\includegraphics[scale = 0.2]{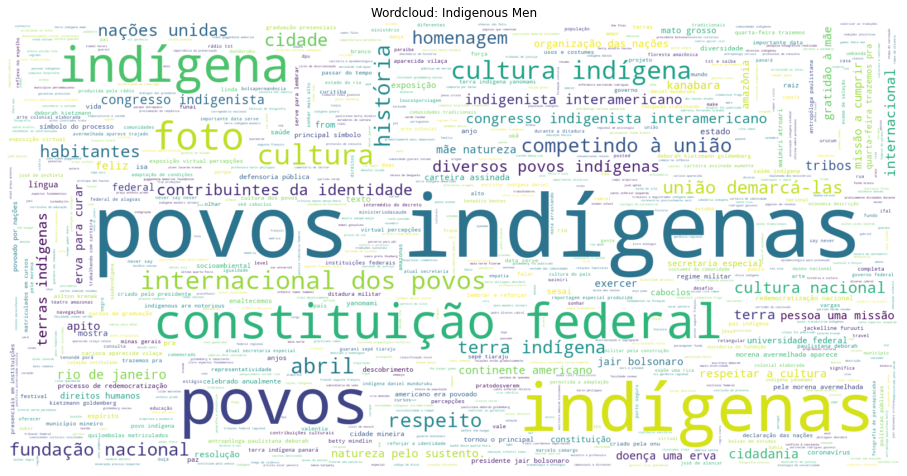} }}
    
    \subfloat[Word cloud related to white women.]{{\includegraphics[scale = 0.2]{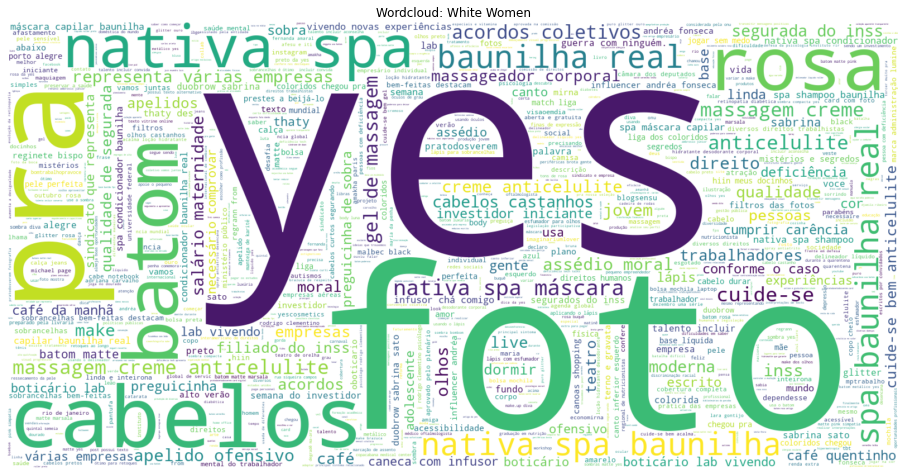} }}\hspace{0.05cm}
    \subfloat[Word cloud related to white men.]{{\includegraphics[scale = 0.2]{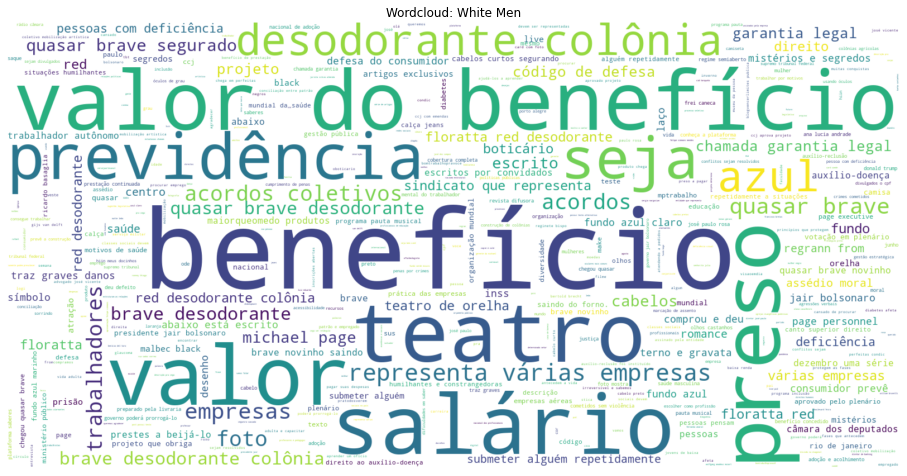} }}
    
    \caption{Word clouds showing the most frequent words associated with women and men from different ethnicities.} 
    \label{fig:stereotype}
\end{figure}

\begin{figure}[h]
    \centering
    \subfloat[Word cloud related to thin women.]{{\includegraphics[scale = 0.2]{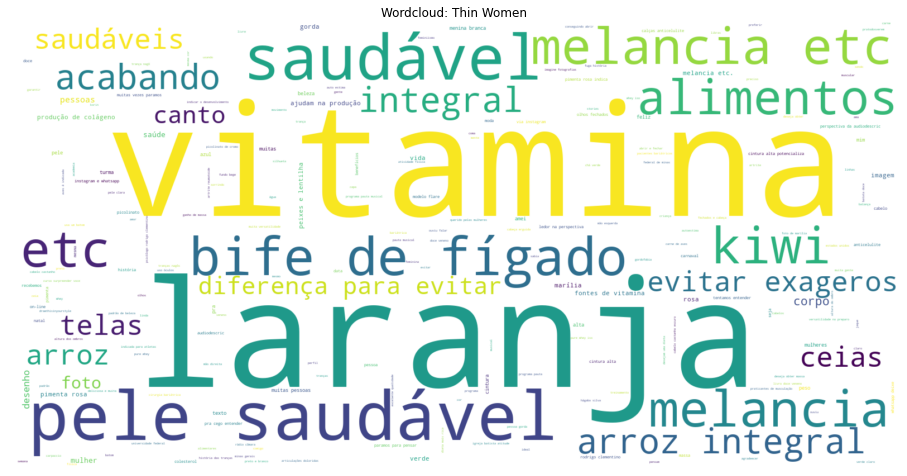} }}\hspace{0.05cm}
    \subfloat[Word cloud related to fat women.]{{\includegraphics[scale = 0.2]{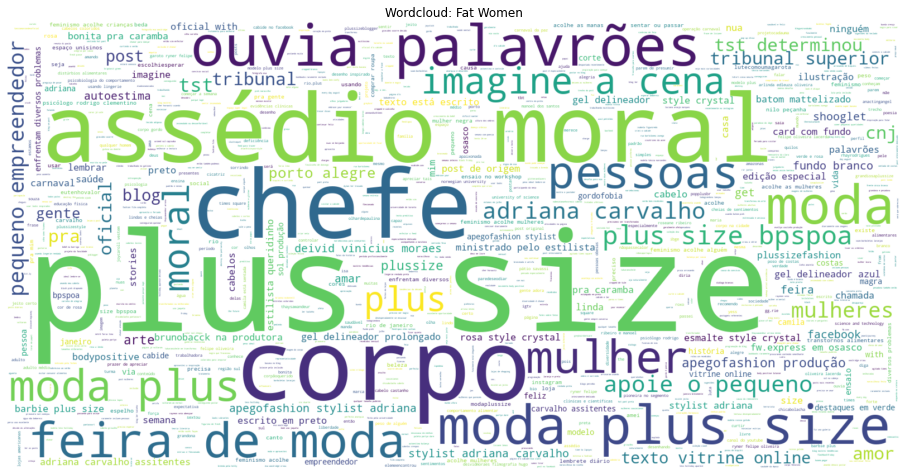} }}
   
    \caption{Word clouds showing the most frequent words associated with fat and thin women. Thin women are associated with vitamins, fruits, and a healthy lifestyle in general.  In contrast, fat women are associated with plus-size style, boss, mental harassment, and swear words.}
    \label{fig:stereotype_wordcloud}
\end{figure}

We collected the data from public posts on Instagram. Thus the data is susceptible to the bias of its algorithm and stereotypes. We conducted an initial analysis of the bias within our dataset. Figure~\ref{fig:stereotype} shows that women are frequently associated with beauty, cosmetic products, and domestic violence. Moreover, black women co-occur more often with terms such as ``racism'', ``discrimination'', ``prejudice'' and ``consciousness'', whereas white women appear with ``spa'', ``hair'' and ``lipstick'', and indigenous women are mostly associated with beauty products. Similarly, black men frequently appear together with the terms ``Zumbi dos Palmares'', ``consciousness'', ``racism'', ``United States'' and ``justice'', while white men are associated with ``theare'', ``wage'', ``benefit'' and ``social security''. In addition, Table~\ref{table:fat_thin_words} shows that women are more frequently associated with physical words (\textit{e.g.}, thin, fat); still, fat people appear more frequently than thin people. Figure~\ref{fig:stereotype_wordcloud} illustrates that fat women are also related to swearing words, ``mental harassment'', ``boss'', while thin women are associated with ``vitamin'', fruits, ``healthy skin''. To sum up, depending on the usage of this dataset, future users may take these aspects into account.

\begin{table}[h]
\caption{This table illustrates how physical characteristics are more related to woman. In addition, ``fat people'' occur more frequently than ``thin people''; these scenarios should be taken into account to avoid biases on models.}
\small
\label{table:fat_thin_words}
\begin{tabular}{lcc}
\toprule
\textbf{words} & \textbf{\#occurrences} & \textbf{ranking} \\ \midrule
gordo          & 467                     & 18,145            \\
gordos         & 26                      & 62,002            \\
gorda          & 806                     & 8,579             \\
gordas         & 176                     & 27,263            \\
magro          & 286                     & 20,161            \\
magros         & 20                      & 68,425            \\
magra          & 410                     & 17,853            \\
magras         & 96                      & 30,195 \\
\bottomrule
\end{tabular}%
\end{table}

\subsection*{Distribution}

\subsubsection*{Will the dataset be distributed to third parties outside of the entity on behalf of which the dataset was created?}
We released the dataset under license CC BY-NC-SA 3.0. We request that the ones who use this dataset cite this paper. Commercial use of this dataset is not permitted.

\subsubsection*{How will the dataset be distributed (e.g., tarball on website, API, GitHub)? Does the dataset have a digital object identifier? }
The dataset will be available (\textit{upon acceptance of this paper}).

\subsubsection*{When will the dataset be distributed? }
\textit{\#PraCegoVer} dataset will initially release \textit{upon acceptance of this paper}, and new versions will be released from time to time.

\subsubsection*{Will the dataset be distributed under a copyright or other intellectual property license, and/or under applicable terms of use?}
The dataset is released under CC BY-NC-SA 3.0. We request that the ones who use this dataset cite this paper. Commercial use of this dataset is not permitted.

\subsubsection*{Do any export controls or other regulatory restrictions apply to the dataset or to individual instances?}
We will share the dataset by filling in a form that identifies who download the data.

\subsection*{Maintenance}

\subsubsection*{Who is supporting/hosting/maintaining the dataset? }
\textit{\#PraCegoVer} dataset is maintained by G.O.S.. All comments or requests can be sent to the email address \texttt{g194760@dac.unicamp.br}. The dataset is hosted by Institute of Computing, University of Campinas.

\subsubsection*{How can the curator of the dataset be contacted?} 
All comments or requests can be sent to G.O.S. through the email address \texttt{g194760@dac.unicamp.br}.

\subsubsection*{Will the dataset be updated?}
\textit{\#PraCegoVer} dataset is continuously updated, and the authors of this dataset will release new versions with new or deleted instances on the GitHub repository. In addition, we will release notes on the \textit{\#PraCegoVer} repository (\url{https://github.com/larocs/PraCegoVer}) with the updates.

\subsubsection*{Will older versions of the dataset continue to be maintained?}
We will keep track of old dataset versions. Thus, they will be available for download.

\subsubsection*{If others want to contribute to the dataset, is there a mechanism for them to do so?}
Contributors shall contact the maintainer, Gabriel O. dos Santos, by email (\texttt{g194760@dac.unicamp.br}).

\end{document}